\documentclass[10pt,twocolumn,letterpaper]{article}

\usepackage[dvipsnames,usenames]{xcolor}
\usepackage{tikz}
\usetikzlibrary{positioning, calc}
\usepackage{iccv}
\usepackage{times}
\usepackage{epsfig}
\usepackage{graphicx}
\usepackage{amsmath}
\usepackage{amssymb}
\usepackage{caption}
\usepackage{pifont}
\usepackage{microtype}
\usepackage{subfigure}
\usepackage{booktabs} %

\usepackage{mathtools}
\usepackage{amsthm}
\usepackage{multicol}
\usepackage{multirow}
\usepackage{graphbox}
\usepackage{adjustbox}
\usepackage[accsupp]{axessibility}

\theoremstyle{plain}

\theoremstyle{definition}

\theoremstyle{remark}

\usepackage[textsize=tiny]{todonotes}

\newlength\savewidth

\def\figMNIST#1{
\begin{figure}[#1]

    \centering
    \begin{tabular}{@{}c@{\,\,}c@{\,\,}@{}}
    \large\rotatebox{90}{\ \quad }  & \includegraphics[width=0.18\linewidth]{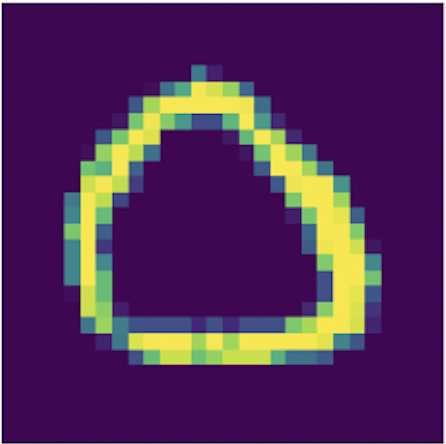} \includegraphics[width=0.18\linewidth]{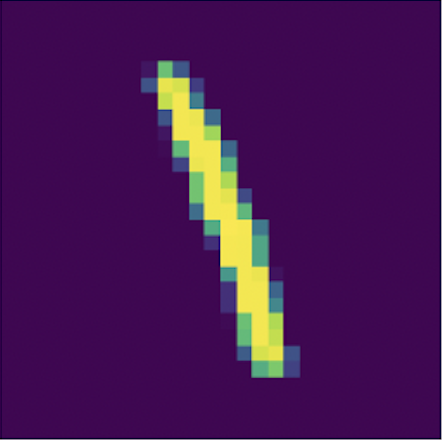}\includegraphics[width=0.18\linewidth]{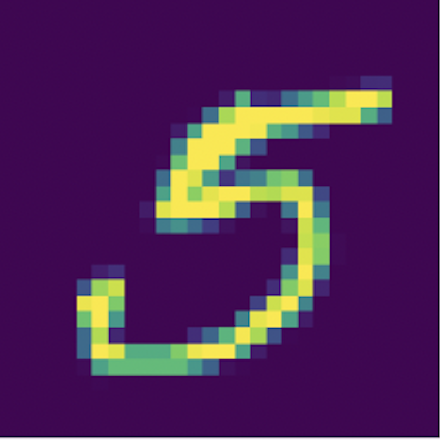}\includegraphics[width=0.18\linewidth]{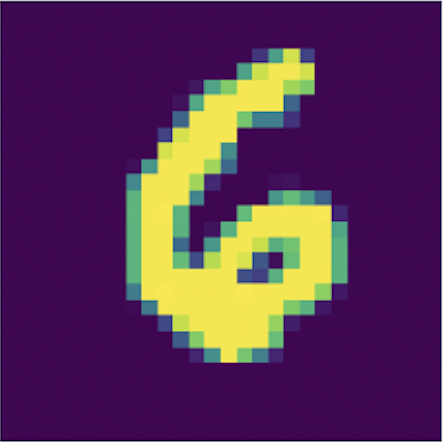}\includegraphics[width=0.18\linewidth]{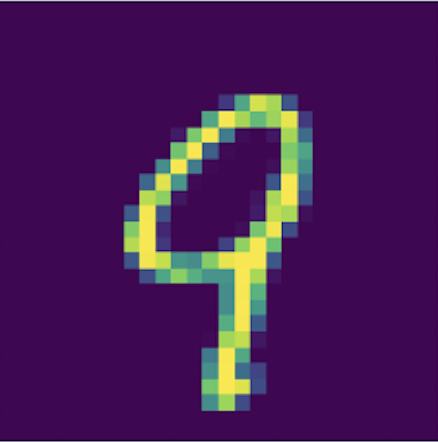} \\
    \large\rotatebox{90}{\ \ \ Ours }  & 
    \includegraphics[width=0.18\linewidth]{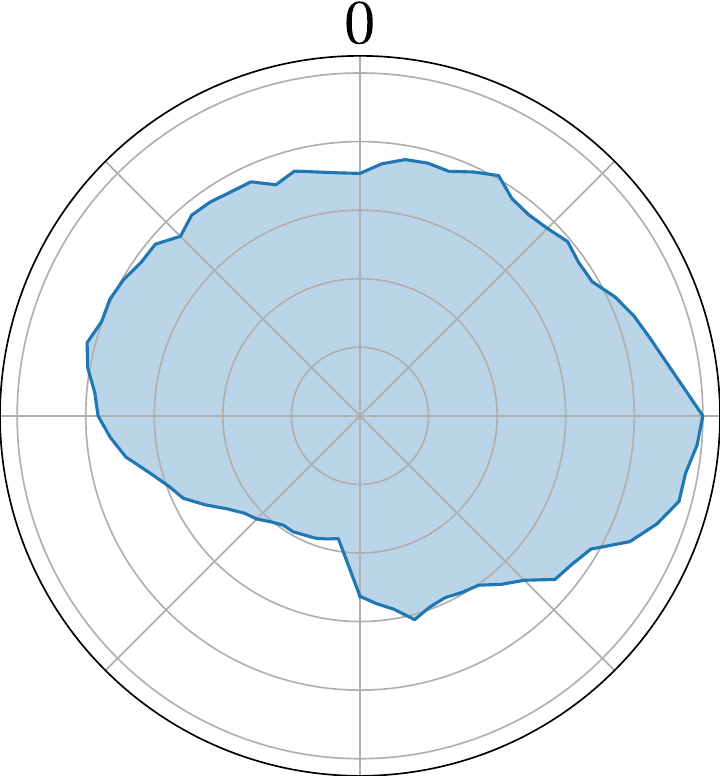} \includegraphics[width=0.18\linewidth]{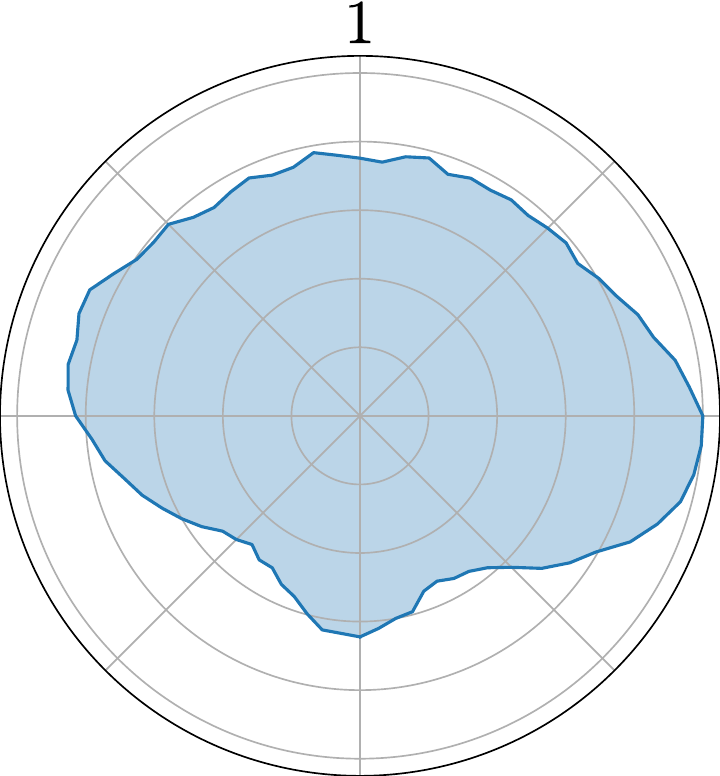}\includegraphics[width=0.18\linewidth]{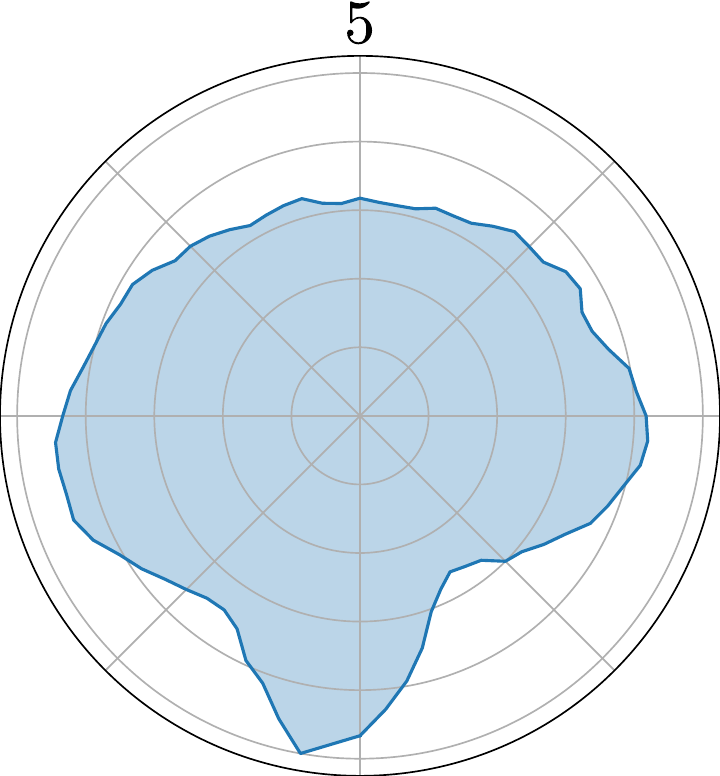}\includegraphics[width=0.18\linewidth]{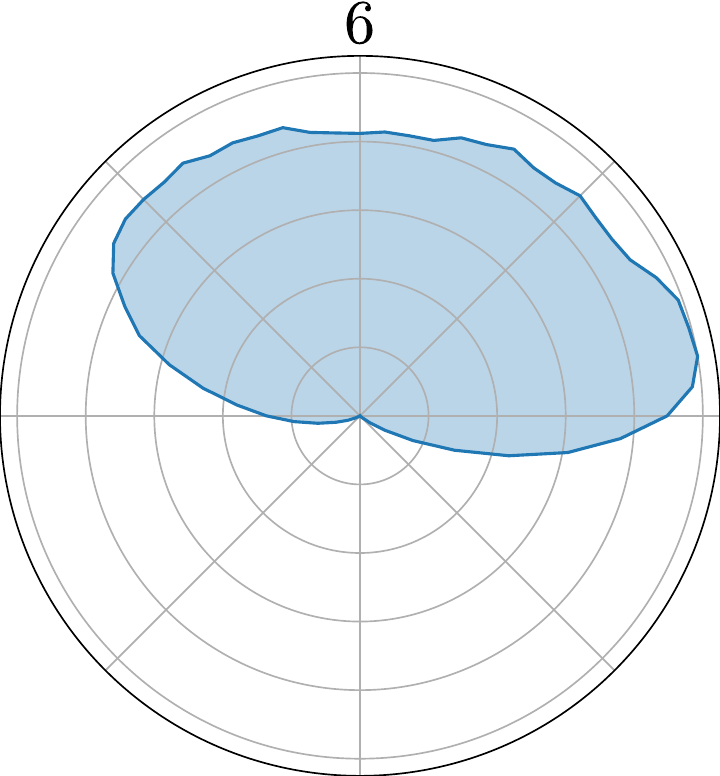}\includegraphics[width=0.18\linewidth]{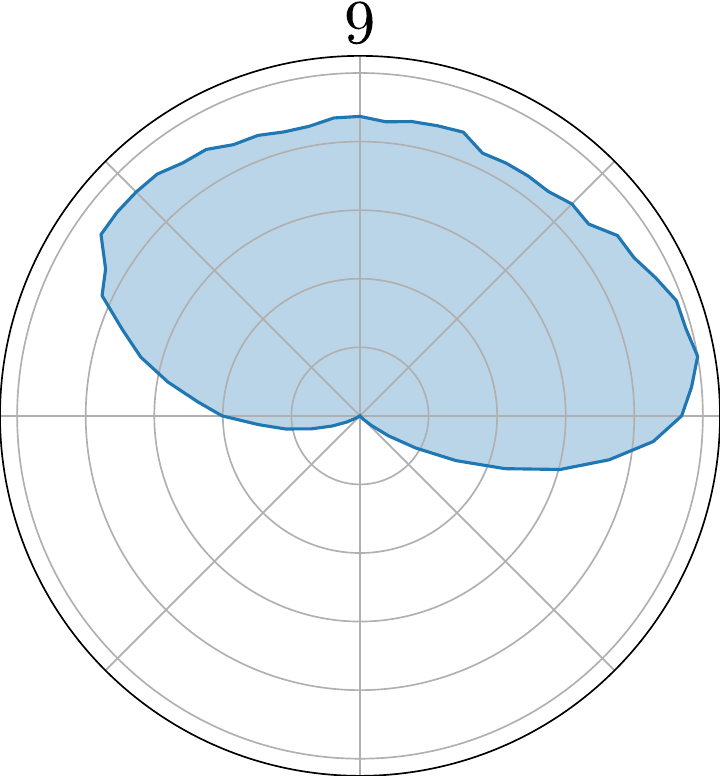}\\
    \large\rotatebox{90}{\  Augerino }  &
    \includegraphics[width=0.18\linewidth]{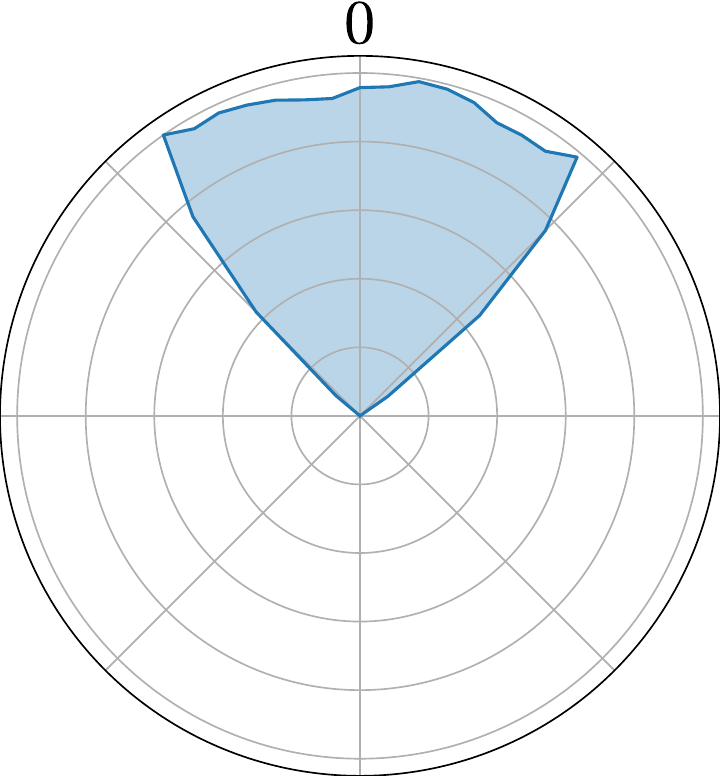} \includegraphics[width=0.18\linewidth]{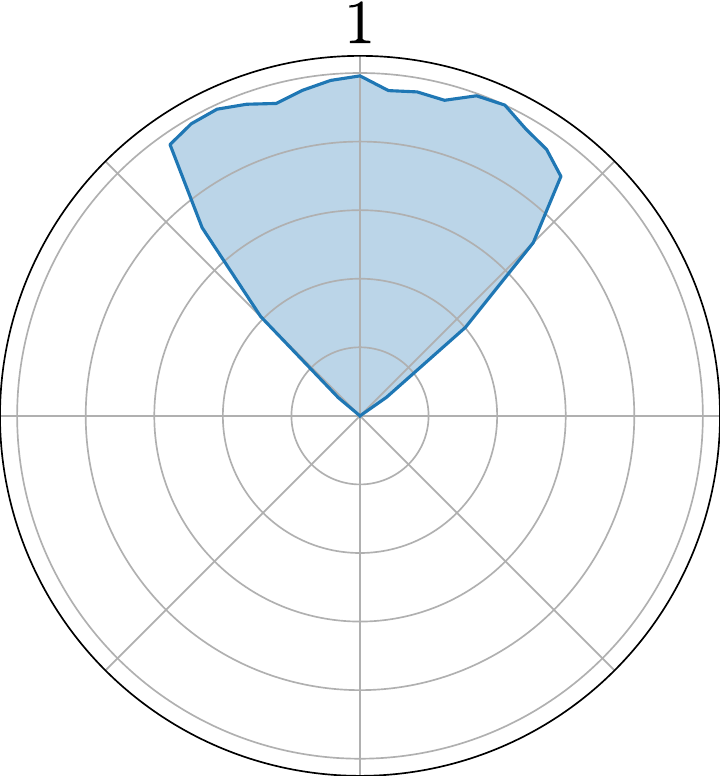}\includegraphics[width=0.18\linewidth]{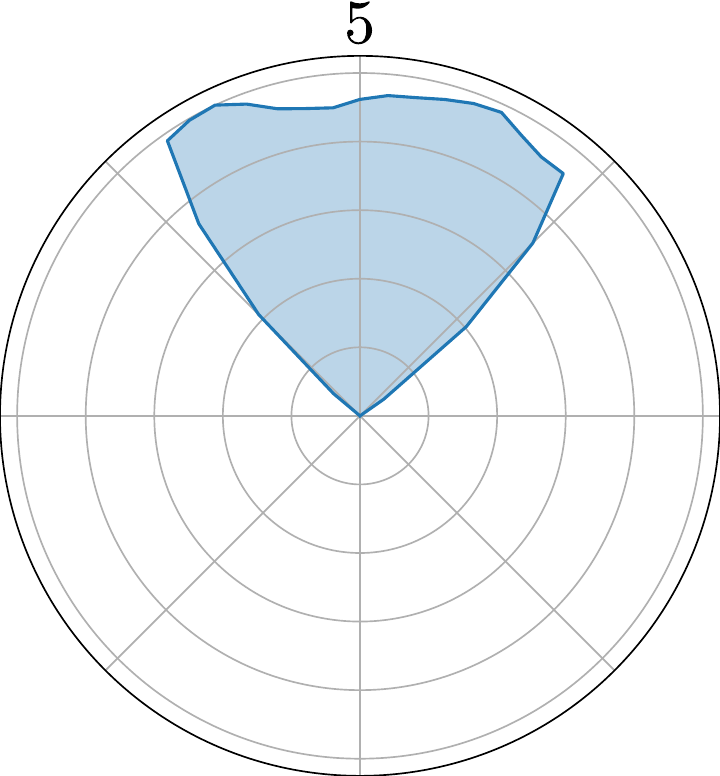}\includegraphics[width=0.18\linewidth]{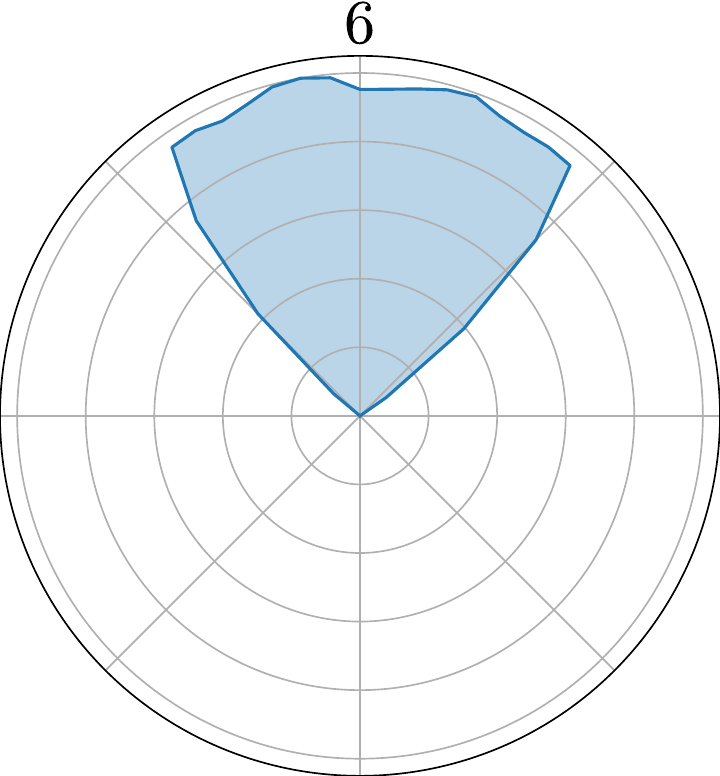}\includegraphics[width=0.18\linewidth]{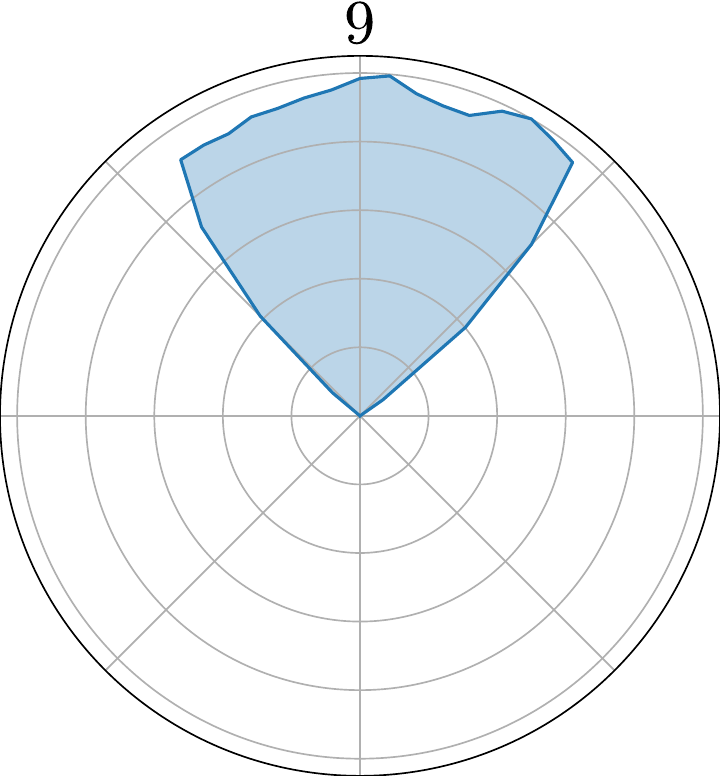} \\
    \end{tabular}
    \includegraphics[width=0.95\linewidth]{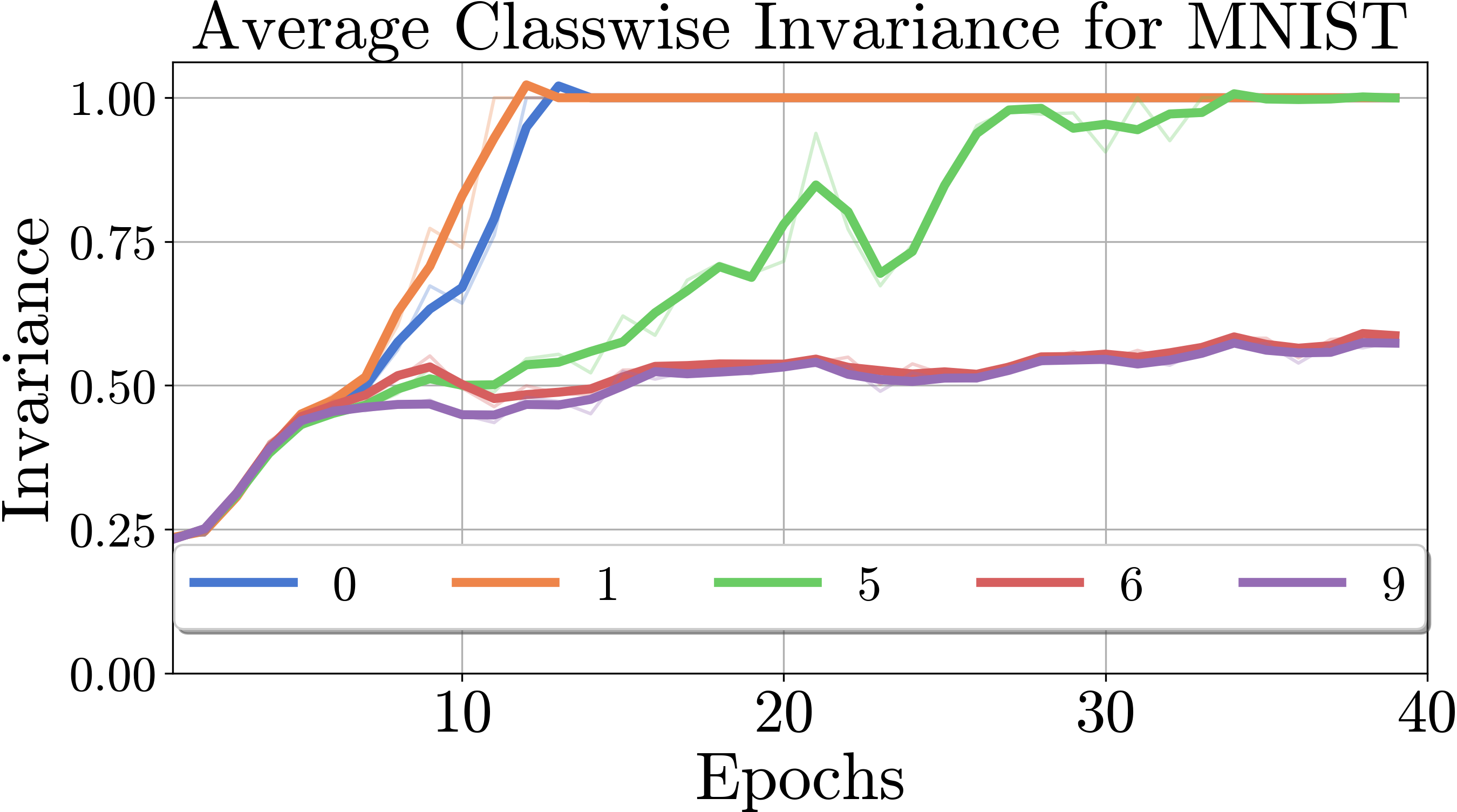} \\
    \caption{Our method learns flexible instance-wise augmentation distributions. We illustrate learned invariance for a subset of MNIST digits ($0$,$1$,$5$,$6$,$9$). The classes $0$,$1$,$5$ can be learned with full invariance, whereas $6$ and $9$ require partial invariance ($\pm 90^{\circ}$). Our model (top) can learn the correct instance-dependent range, whereas Augerino (middle) instead learns a much narrower shared invariance for all classes. (bottom) A plot of the classwise learned rotational invariance for our model over time. Classes 0, 1, and 5 achieve close to full rotational invariance, whereas 6 and 9 achieve close to $\pm 90^{\circ}$ rotational invariance.}
    \label{fig:mnist}
    \vspace{-5mm}
\end{figure}
}

\def\figInvarianceTransferClass#1{
\begin{figure}[#1]
    \centering
    \includegraphics[width=0.9\linewidth]{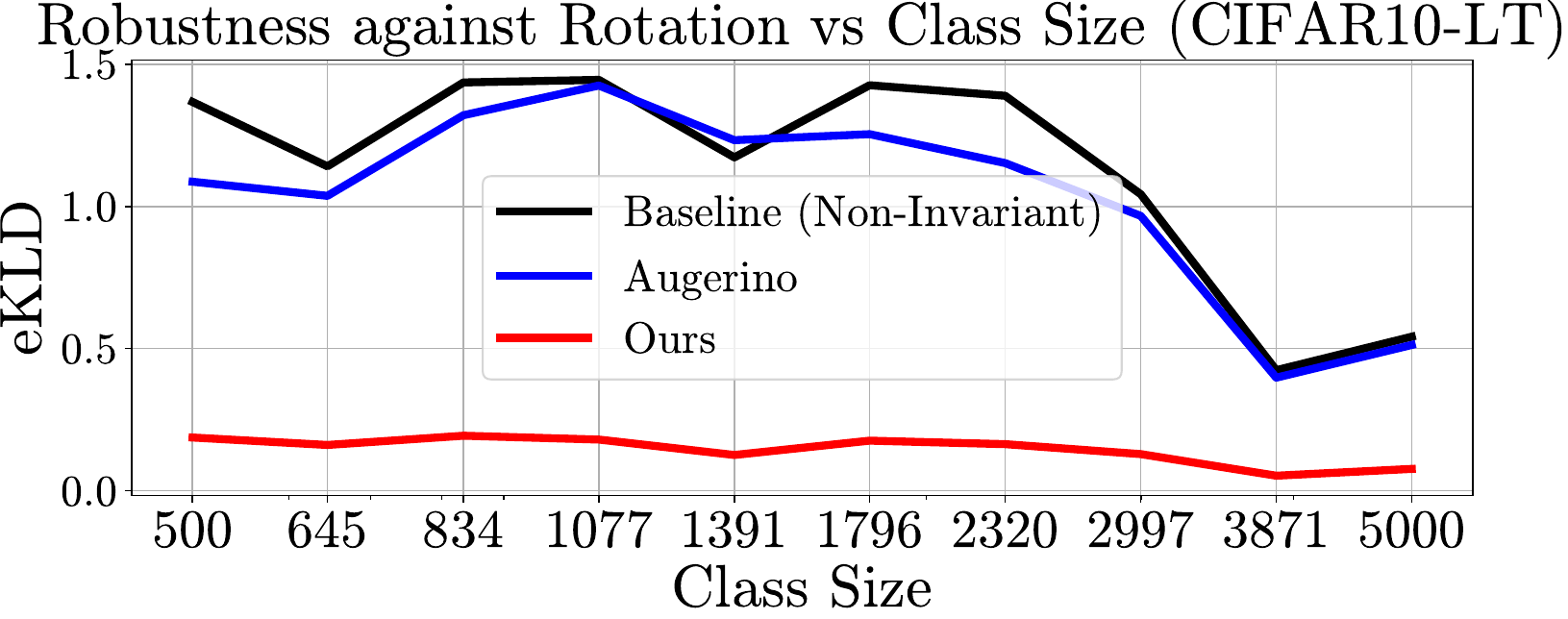} \\
    \includegraphics[width=0.9\linewidth]{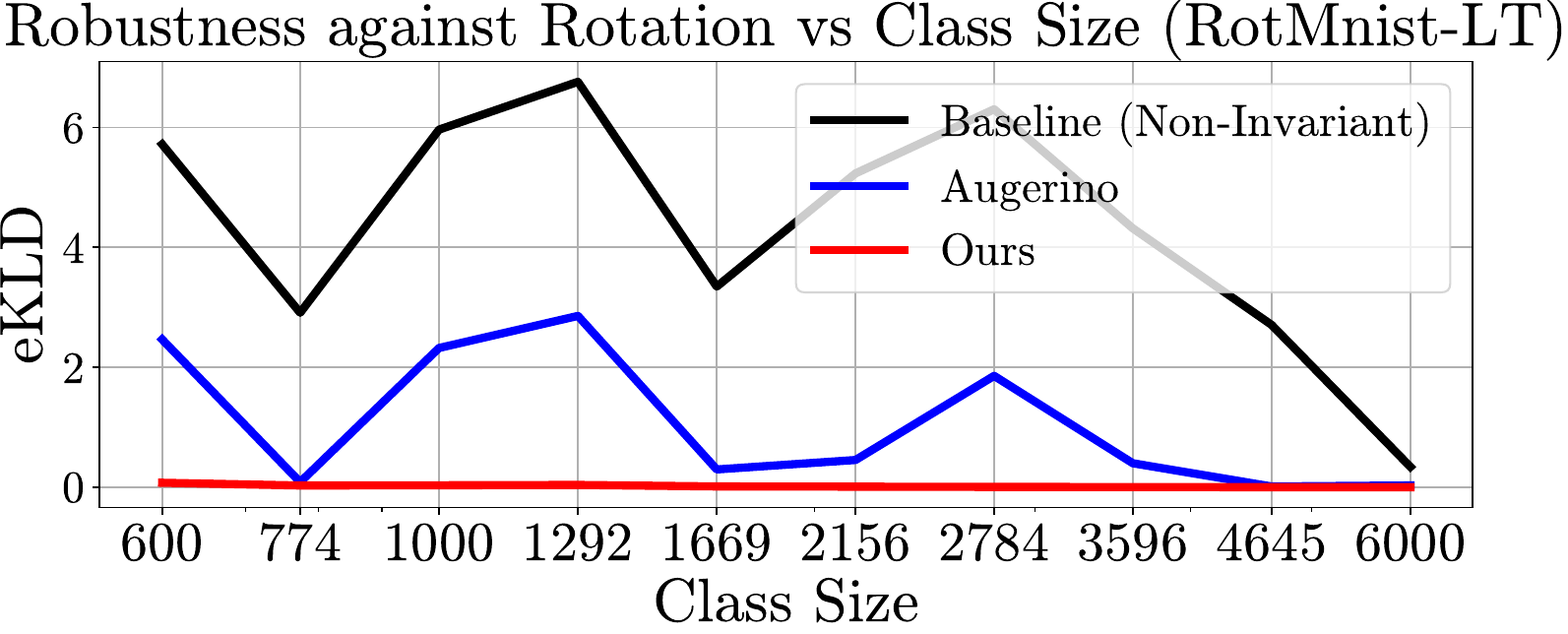}
    \caption{Invariance transfer from head classes to tail classes in imbalanced classification. We follow Zhou~\etal~\cite{git} (Fig 3) and plot the expected KL-divergence under image rotations for RotMNIST-LT and CIFAR10-LT (lower is better). RotMNIST-LT is a long-tail version of the MNIST dataset where each image has been randomly rotated. As Zhou~\etal~\cite{git} shows, neural networks learn rotational invariance for head classes (indicated by low eKLD) but fail to transfer this invariance to tail classes. This problem persists for Augerino to a lesser extent. In contrast, our method successfully transfers invariance across classes. This effect is even more pronounced for CIFAR10-LT $(\pm 10^{\circ}$ rotations)}
    \label{fig:transfer_class}
    \vspace{-5mm}
\end{figure}
}

\def\figGraph#1{
\begin{figure}[#1]
    \centering

    \begin{tikzpicture}
        \node[anchor=south west,inner sep=0] (image) at (0,0) {\includegraphics[width=0.8\linewidth]{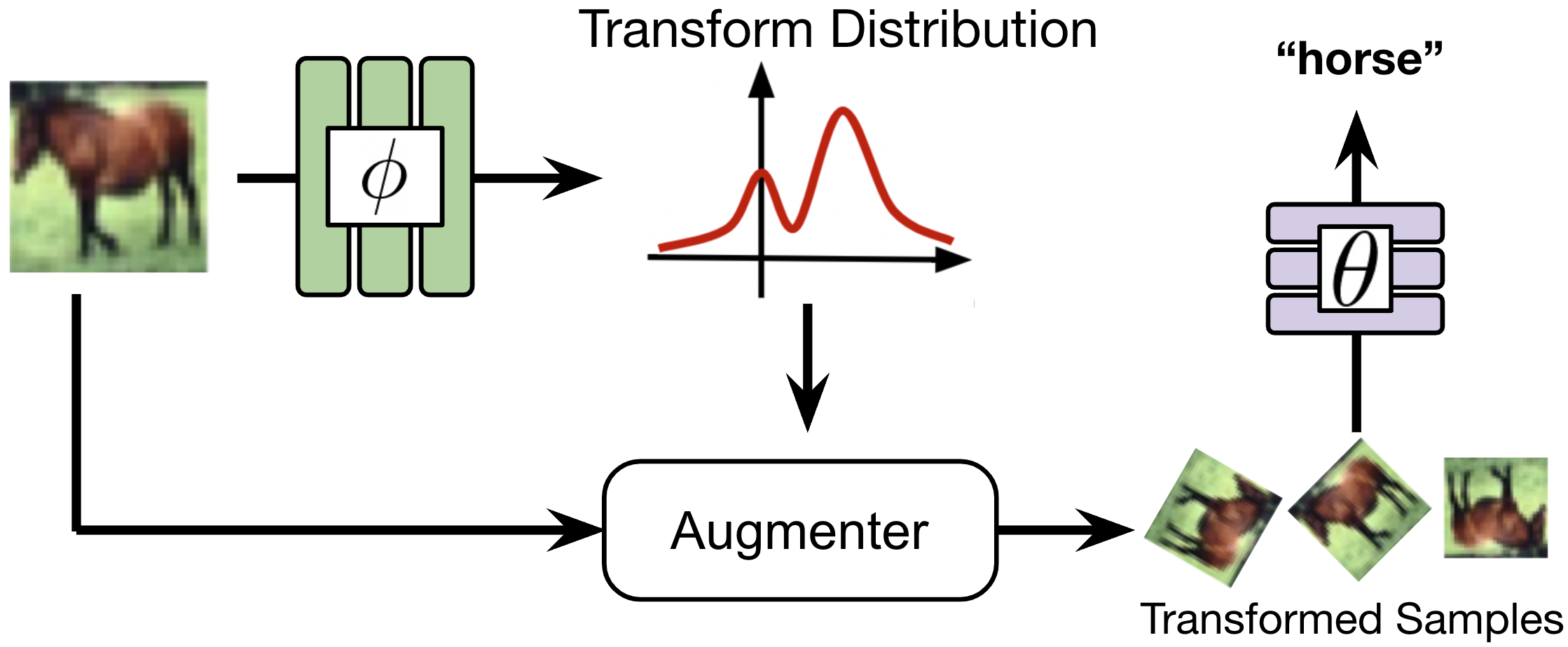}};
        \node[anchor=west,yshift=-0.2cm] at (image.north west) {Input};
    \end{tikzpicture}
    \caption{Our image classification pipeline. The normalizing flow model predicts a distribution over image transformations. Samples from this distribution are passed to a differentiable augmented, which transforms the input image into a set of augmented images. The images are passed to a classifier, and predictions are averaged. Crucially, the transform distribution $g_\phi$ can generalize across classes and datasets.}
    \label{fig:graph}
    \vspace{-5mm}
\end{figure}
}

\def\figTTRobust#1{
\begin{figure*}[#1]
    \centering
    
    \newlength{\mylength}
    \setlength{\mylength}{0.06\linewidth} %

    \begin{tabular}[c]{@{}ccc@{}}
    \begin{tabular}[c]{@{}c@{}}
    (a) \\
    \includegraphics[width=1.2\mylength]{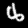}\\
    \includegraphics[width=1.2\mylength]{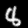}\\
    \includegraphics[width=1.2\mylength]{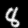}\\
    \includegraphics[width=1.2\mylength]{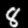}\\
    \includegraphics[width=1.2\mylength]{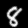}
    \end{tabular} &
    \begin{tabular}[c]{@{}c@{}}
    \begin{tabular}[c]{@{}c@{}c@{}}
    \begin{tabular}[c]{@{}c@{}}
    (b) \\
    \includegraphics[trim=0 0 70 0,clip,height=3\mylength]{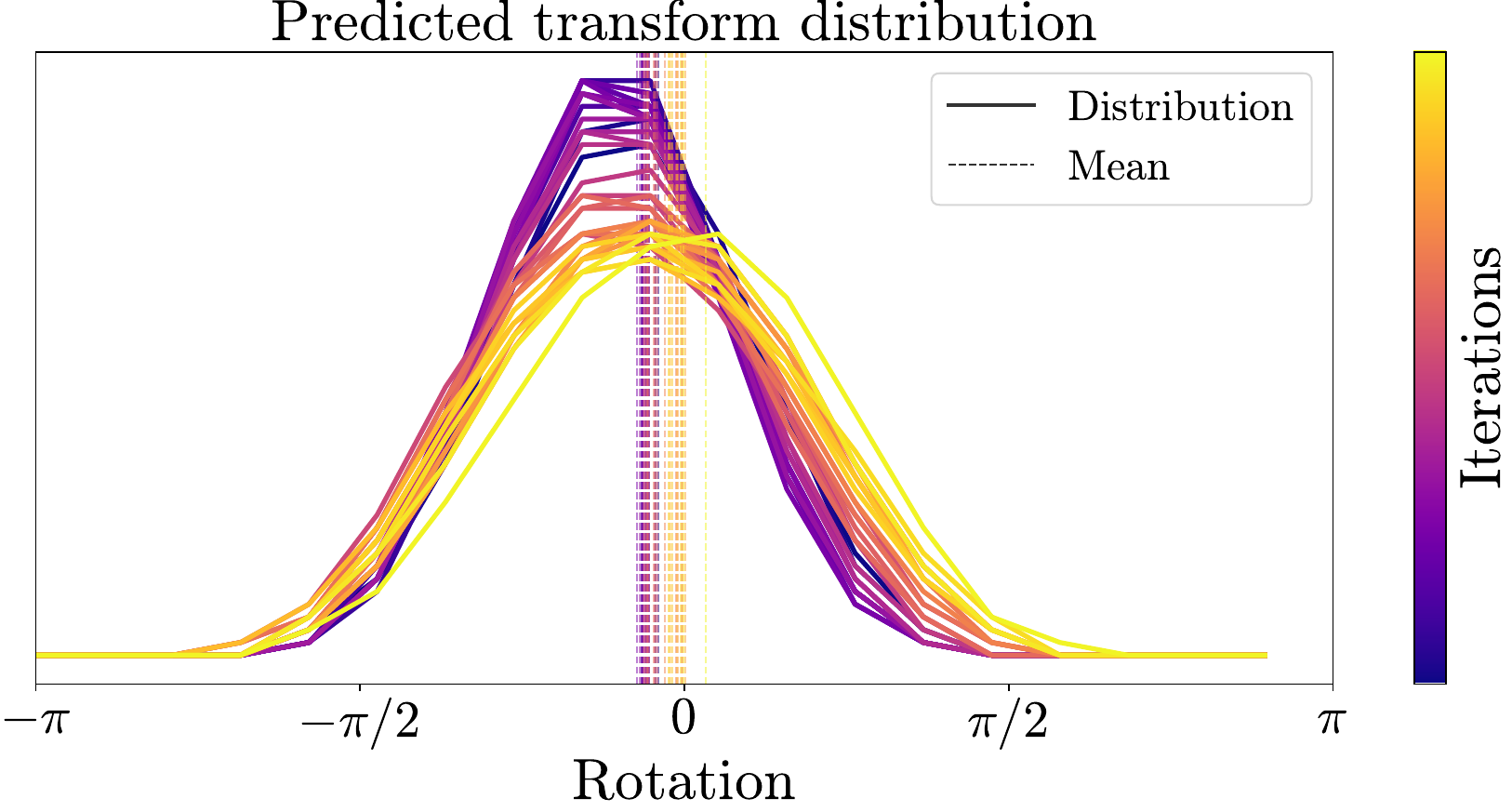} \end{tabular} &
    \begin{tabular}[c]{@{}c@{}}
    (c) \\
    \includegraphics[height=3\mylength]{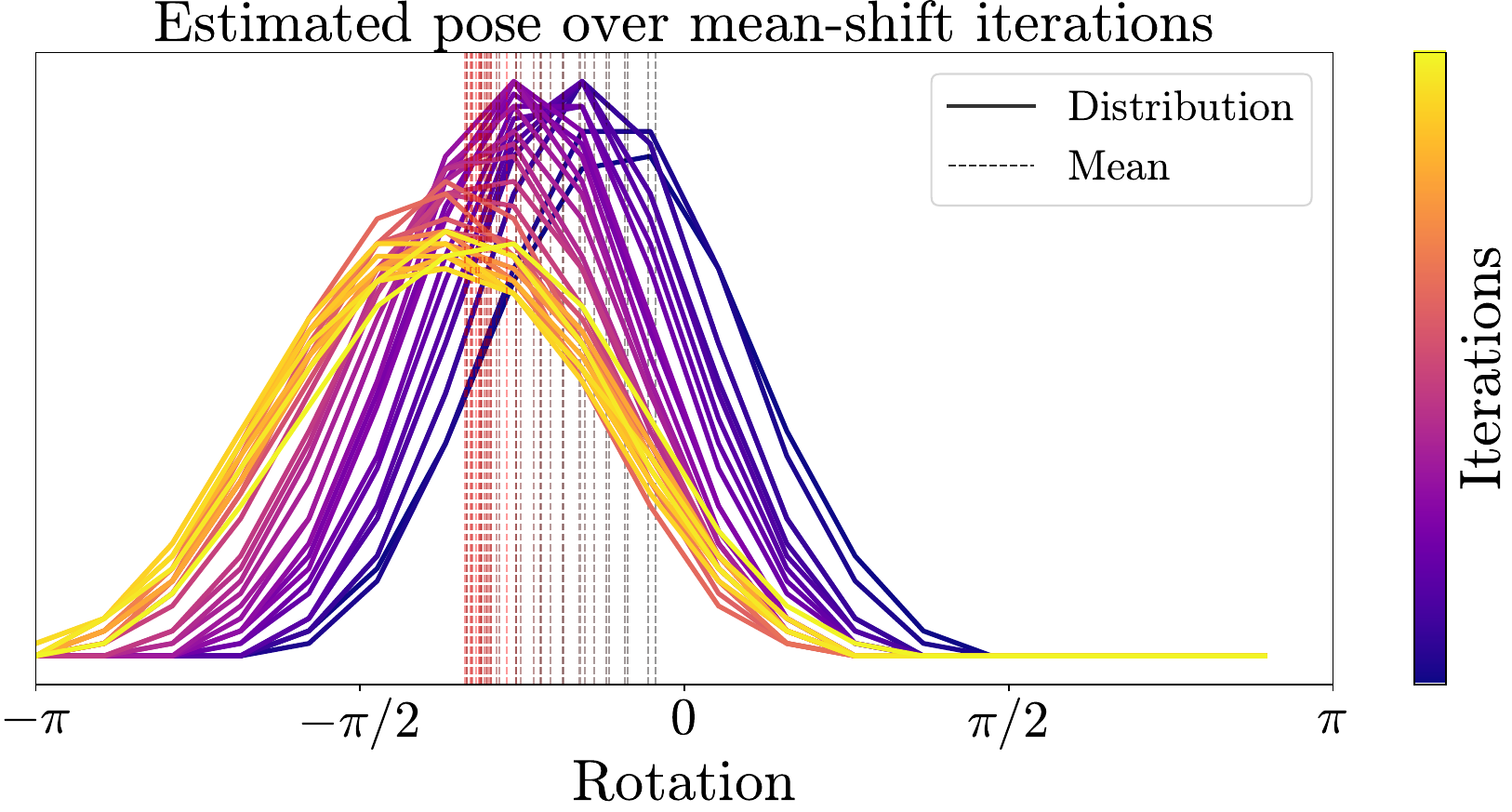} \end{tabular} \\
    \end{tabular} \\
    \begin{tabular}[c]{@{}cc@{}}
    \begin{tabular}[c]{@{}c@{}}
    (d) \\
    \includegraphics[height=3.5\mylength]{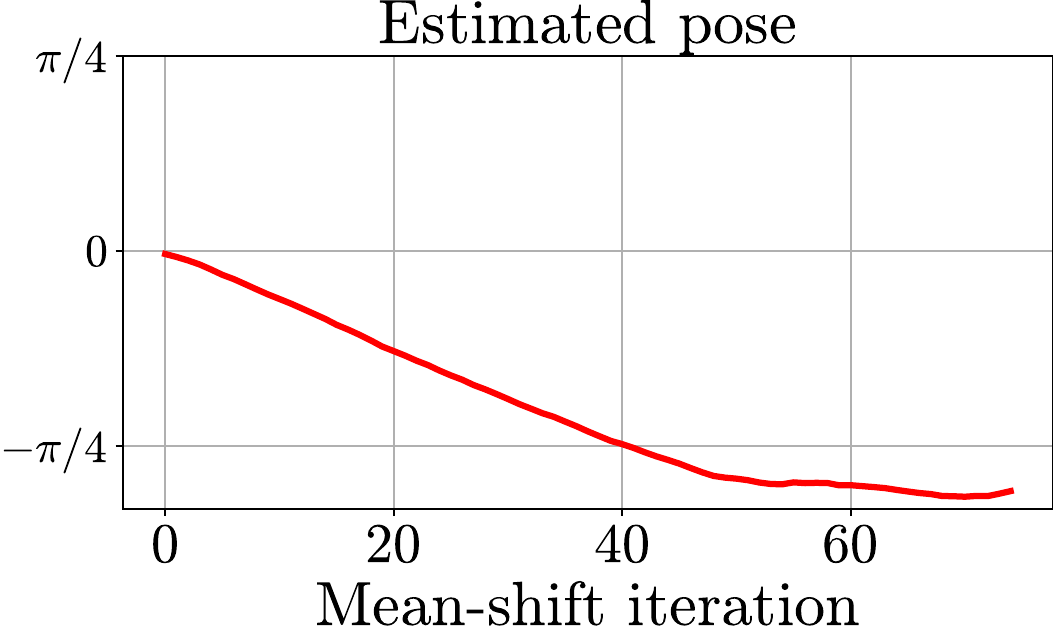} \end{tabular} & \begin{tabular}[c]{@{}c@{}}
    (e) \\
    \begin{tabular}{c@{}c@{}c@{}c}
    \includegraphics[width=\mylength]{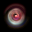} &\includegraphics[width=\mylength]{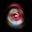} &\includegraphics[width=\mylength]{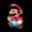}&
    \includegraphics[width=\mylength]{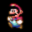}\\
    \includegraphics[width=\mylength]{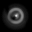} &\includegraphics[width=\mylength]{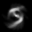} &\includegraphics[width=\mylength]{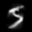}&
    \includegraphics[width=\mylength]{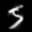}\\
    \includegraphics[width=\mylength]{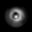} &\includegraphics[width=\mylength]{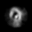} &\includegraphics[width=\mylength]{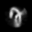}&
    \includegraphics[width=\mylength]{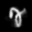}\\
    \end{tabular} \\
    \end{tabular}
    \end{tabular} 
    \end{tabular} &
    \begin{tabular}{@{}c@{}}
    (f)\\
    \begin{tabular}{@{}c@{}c@{}}
    \includegraphics[width=0.6\mylength]{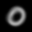}& \includegraphics[width=0.6\mylength]{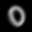}\\ \includegraphics[width=0.6\mylength]{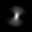}& \includegraphics[width=0.6\mylength]{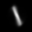}\\ \includegraphics[width=0.6\mylength]{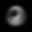}& \includegraphics[width=0.6\mylength]{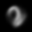}\\ \includegraphics[width=0.6\mylength]{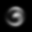}& \includegraphics[width=0.6\mylength]{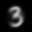}\\ \includegraphics[width=0.6\mylength]{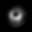}& \includegraphics[width=0.6\mylength]{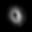} \\ \includegraphics[width=0.6\mylength]{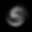}& \includegraphics[width=0.6\mylength]{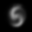}\\ \includegraphics[width=0.6\mylength]{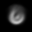}& \includegraphics[width=0.6\mylength]{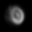}\\ \includegraphics[width=0.6\mylength]{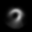}& \includegraphics[width=0.6\mylength]{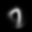}\\ \includegraphics[width=0.6\mylength]{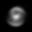}& \includegraphics[width=0.6\mylength]{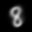}\\ \includegraphics[width=0.6\mylength]{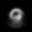}& \includegraphics[width=0.6\mylength]{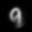} \\

    \end{tabular}
    
    \end{tabular} 
    \end{tabular}
    
    \caption{The conditional augmentation distribution can be used to align an image dataset, discover prototypes similar to congealing \cite{congealing}, and adapt to out-of-distribution poses. \textbf{(a)} Example of the mean shift algorithm aligning a digit belonging to an unseen class. \textbf{(b-d)} Figures showing the modified mean-shift algorithm. For a given input, we repeatedly compute the mean of the conditional transform distribution and perturb the input in that direction, pushing the input close to a local mode. As a result, the mean of the transformation distribution slowly shifts to $0$ while the estimated pose gets closer to the true pose. \textbf{(e)} Demonstration of an augmentation distribution aligning rotated ($\pm 90^\circ$) versions of a single image. We separately apply mean-shift to each rotated image and observe that they converge to the same mode. Unlike \cite{congealing}, there is no joint optimization, and each image is ``aligned'' separately. This alignment also works for MNIST images even though the model has only trained on Mario-Iggy. \textbf{(f)} We apply the model trained on Mario-Iggy to align each class in the MNIST test set, and we make the task more challenging by adding $\pm 45^{\circ}$ rotations to each image. The top row shows the average class image before alignment, and the bottom row shows images after alignment. We successfully discover prototypes for $0,1,3,8,9$, whereas for classes like $4,6$, the model fails due to multiple possible modes. }
    \label{fig:alignment}
    \label{fig:test_time_robust}
\end{figure*}
}

\def\figTeaser#1{
\begin{figure}[#1]
    \centering
    \includegraphics[width=0.9\linewidth]{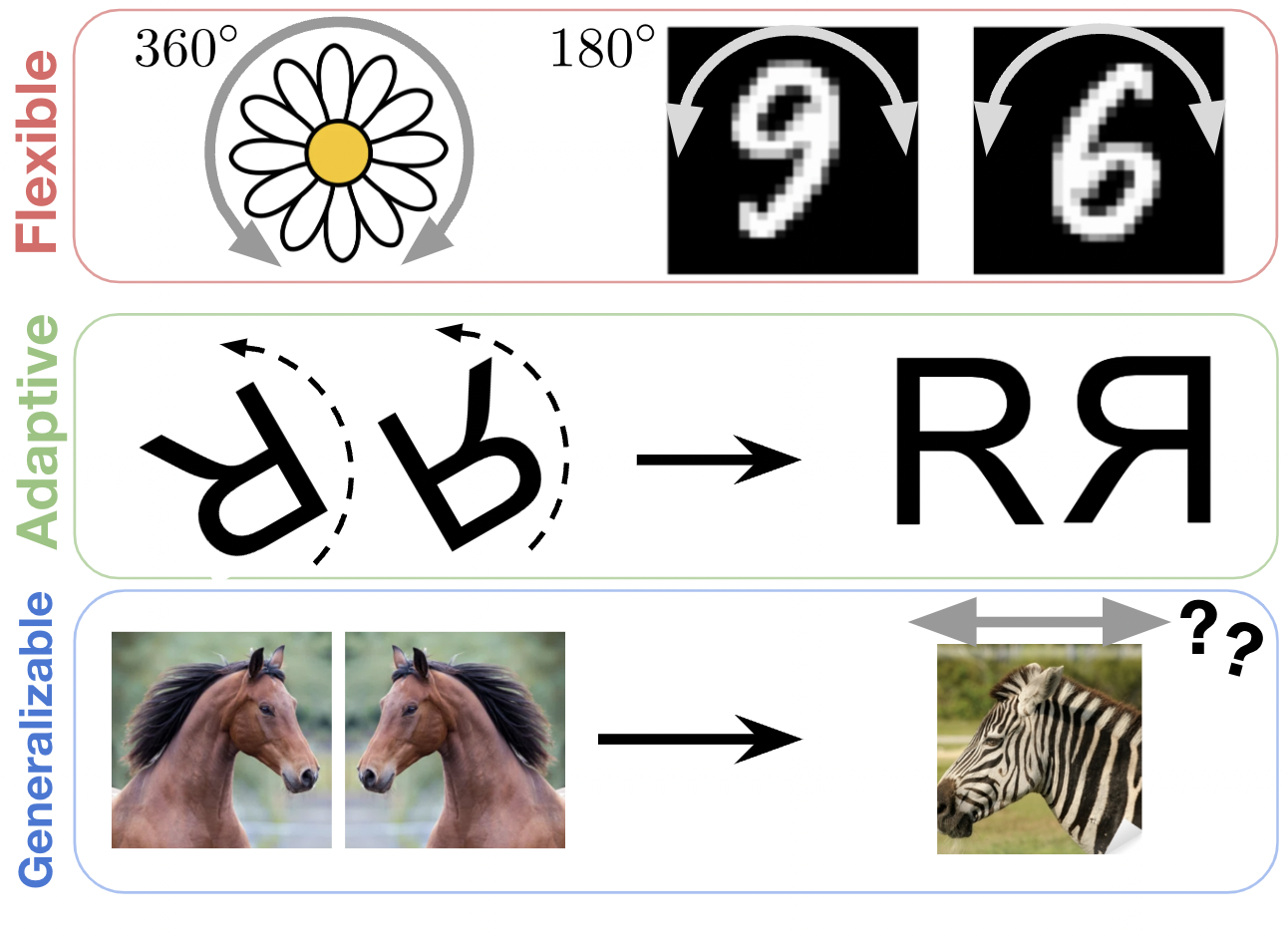}
    \caption{Our goal is to build flexible, adaptive, and generalizable invariances. \textbf{Flexible:} The ideal invariance is flexible and instance-dependent. Different objects in different poses require different degrees of invariance. Too much hurts accuracy, and too little hurts robustness.  \textbf{Adaptive}: The model should adapt to unexpected (out-of-distribution) poses. The figure above shows mental rotation, a process by which humans align unfamiliar objects in unexpected poses to classify them. \textbf{Generalizable}: Knowledge of invariances should generalize from previous experience, e.g., learning bilateral symmetry for horses and transferring it to zebras.}
    \label{fig:teaser}
    \vspace{-5mm}
\end{figure}
}

\def\tabLILA#1{

\begin{table}[#1]
    \centering
        \centering
        \begin{adjustbox}{max width=\linewidth}
        \begin{tabular}{lllll}
            \toprule
                 & \textbf{CIFAR10} & \textbf{FMNIST} & \textbf{MNIST} & \textbf{CIFAR10-LT} \\
            \midrule
            Baseline & 74.1 \small{$\pm$ 0.5} & 89.6 \small{$\pm$ 0.2} & 99.1 \small{$\pm$ 0.02} & 70.8 \small{$\pm$ 0.8} \\
            Augerino & 79.0 \small{$\pm$ 1} & 90.1 \small{$\pm$ 0.1} & 98.3 \small{$\pm$ 0.1} & 63.6 \small{$\pm$ 1.3} \\
            LILA & \underline{84.2} \small{$\pm$ 0.8} & \underline{91.9} \small{$\pm$ 0.2} & \textbf{99.4} \small{$\pm$ 0.02} & \underline{76.4} \small{$\pm$ 0.9} \\
            Ours & \textbf{86.8 \small{$\pm$ 0.4}} & \textbf{92.3 \small{$\pm$ 1.4}}  & \underline{99.2} \small{$\pm$ 0.1} & \textbf{78.1 $\pm$ 1}  \\
            \midrule
            Gain & \textcolor[rgb]{0,0.7,0}{(+2.6)} & \textcolor[rgb]{0,0.7,0}{(+0.4)} & \textcolor[rgb]{0.7,0,0}{(-0.2)} & \textcolor[rgb]{0,0.7,0}{(+1.7)} \\
            \bottomrule
        \end{tabular}
        \end{adjustbox}
        \label{tab:lila}
    \caption{Classification accuracy on the modified ResNet used by LILA \cite{lila}. Numbers for baselines reproduced from \cite{lila}. Our method helps the classifier achieve the highest test accuracy on CIFAR10 and CIFAR10-LT(rho=10). Imbalanced classification is particularly challenging since invariances learned through augmentations do not transfer from head classes to tail classes \cite{git}. We note that our method is complementary to LILA and can be combined in future work. }
\vspace{-3mm}
\end{table}
}

\def\figorth#1{
\begin{figure}[#1]
    \centering
    \begin{tikzpicture}
        \node[inner sep=0pt,anchor=south west] (a) at (0,0) {\includegraphics[height=0.43\linewidth]{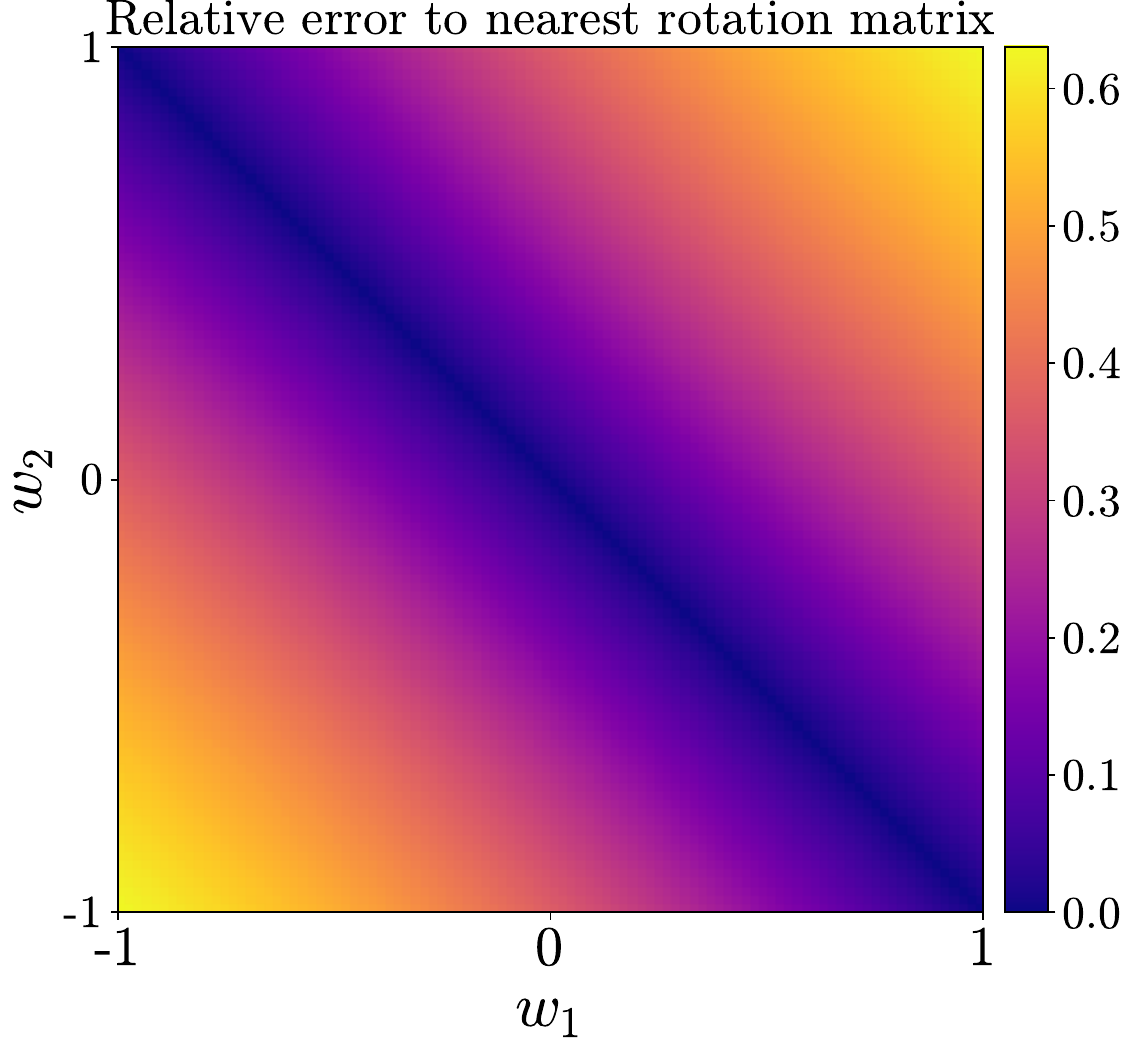}};
        \node[anchor=north west,xshift=-4pt,yshift=-4pt] at (a.north west) {(a)};
        
        \node[inner sep=0pt,anchor=south west] (b) at (a.south east) {\includegraphics[height=0.43\linewidth]{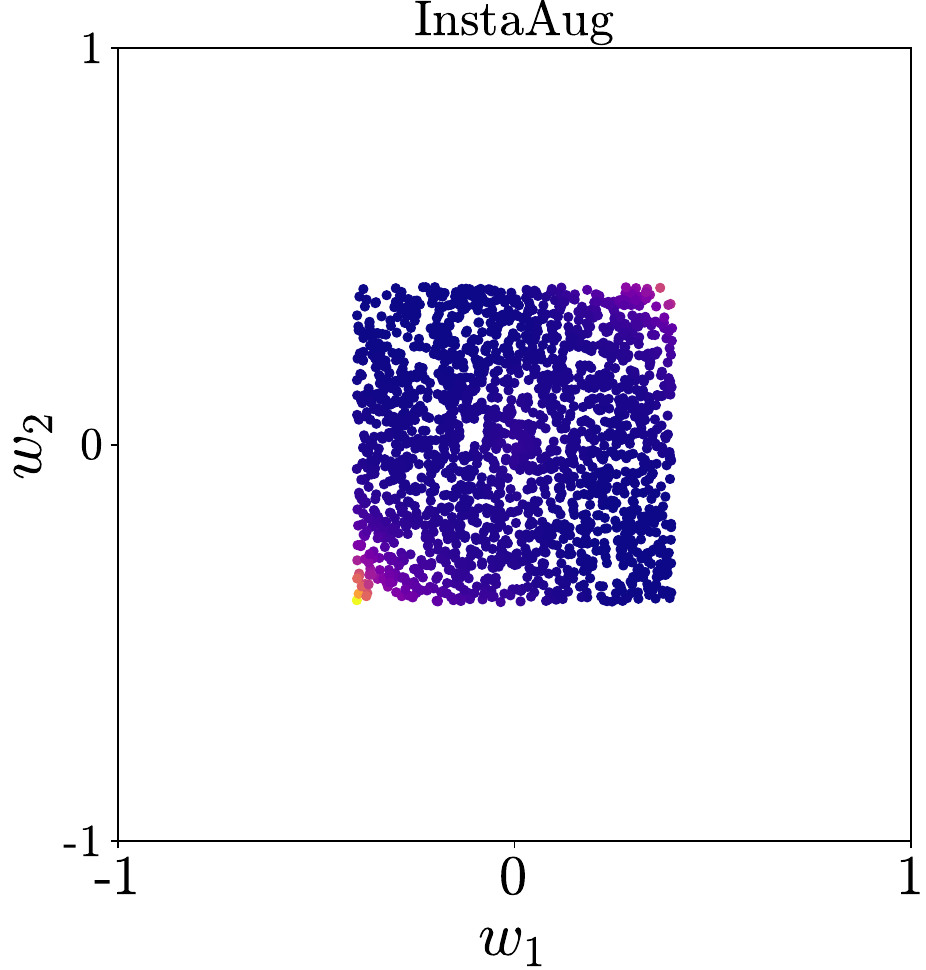}};
        \node[anchor=north west,xshift=-4pt,yshift=-4pt] at (b.north west) {(b)};
        
        \node[inner sep=0pt,anchor=north west] (c) at (a.south west) {\includegraphics[height=0.35\linewidth]{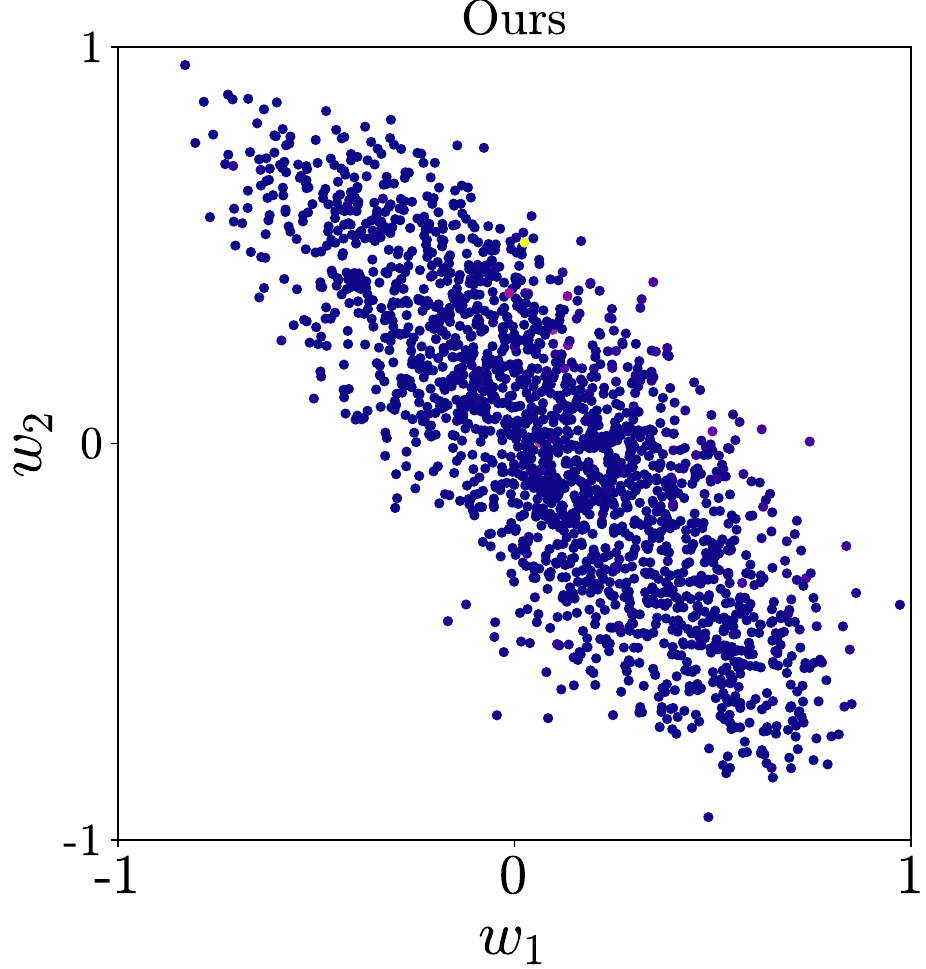}};
        \node[anchor=north west,yshift=10pt] at (c.north west) {(c)};
        
        \node[inner sep=0pt,anchor=north west] (d) at (c.north east) {\includegraphics[trim=460 0 0 0,clip,height=0.35\linewidth]{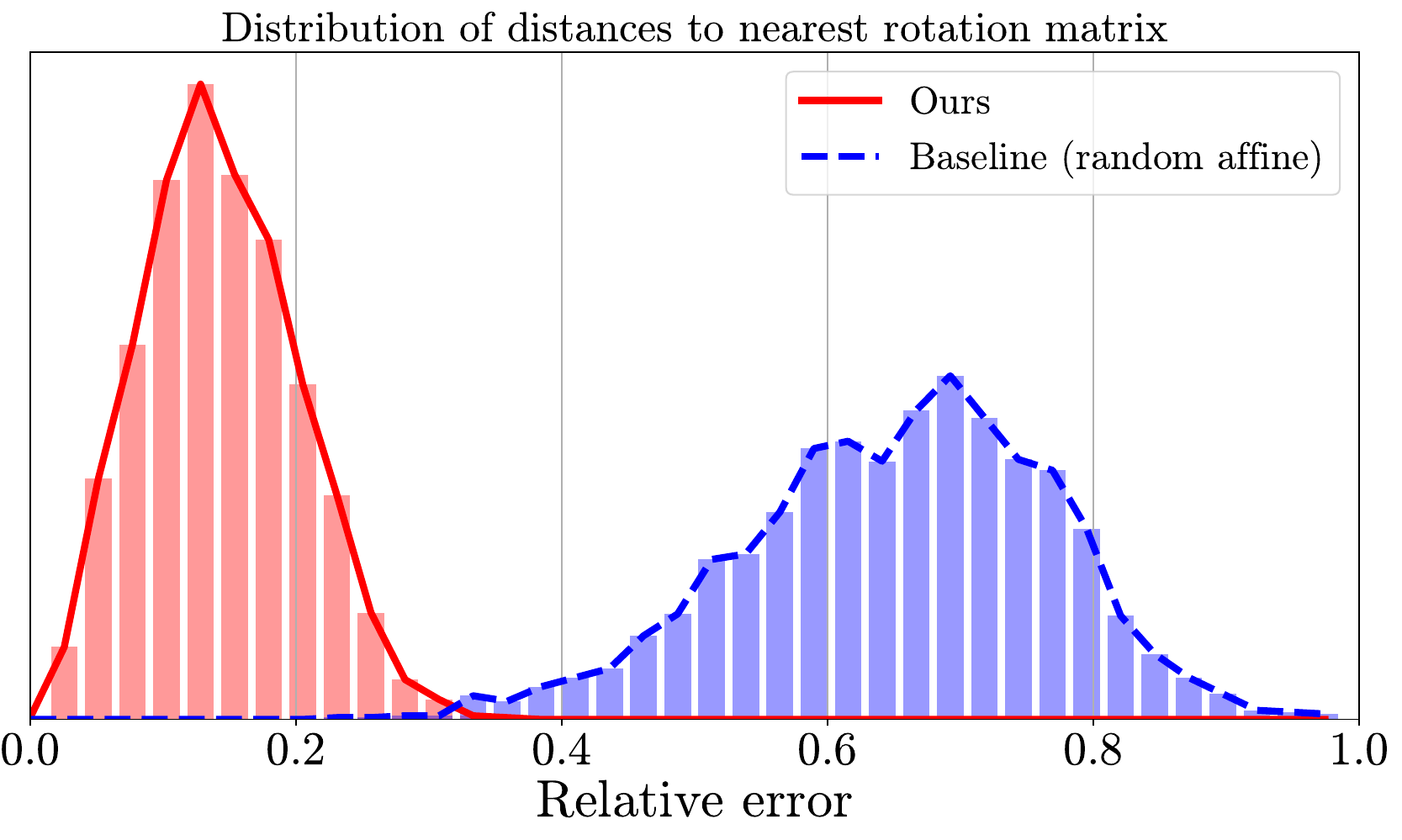}};
        \node[anchor=north west,yshift=10pt] at (d.north west) {(d)};
    \end{tikzpicture}
    \caption{Our model learns the rotation constraint from data, while InstaAug fails to represent non-axis-aligned distributions. The goal of the ``rotation discovery'' task is to learn the joint distribution of affine matrix parameters such that the result is rotation. $(w_1,w_2)$ pairs on diagonal (i.e., $w_1 = -w_2$) correspond to exact rotations and thus incur a small classification loss. \textbf{(a)} Relative error to the nearest rotation matrix. The ideal distribution of augmentations is in the form of a diagonal strip. \textbf{(b)} InstaAug produces a small square as its mean-field parametrization is unable to represent correlations between two parameters. \textbf{(c)} Our model learns to produce samples on the diagonal and learns a much larger range than InstaAug.  \textbf{(d)}  We plot the histogram of relative errors of the produced samples to the nearest rotation matrix. It is much smaller than the random affine baseline. Our model learns the joint distribution and discovers rotations from the full set of affine parameters, while InstaAug fails.}
    \label{fig:orth}
    \vspace{-5mm}
\end{figure}
}

\def\figDistribution#1{
\begin{figure}[#1]
        \centering
            \begin{tabular}{@{}c@{}}
                \includegraphics[width=0.9\linewidth]{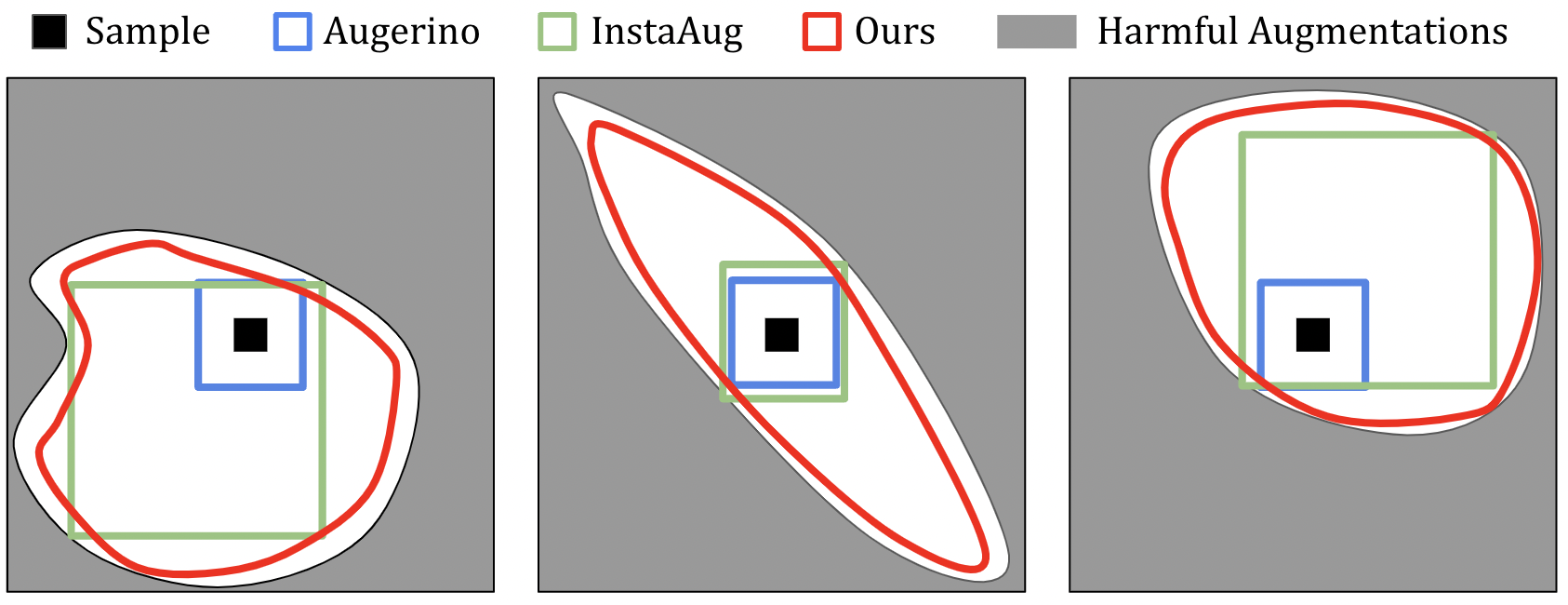}\\
            \end{tabular}
            \begin{tabular}{@{}cc@{}}
                \includegraphics[width=0.2\linewidth,align=c]{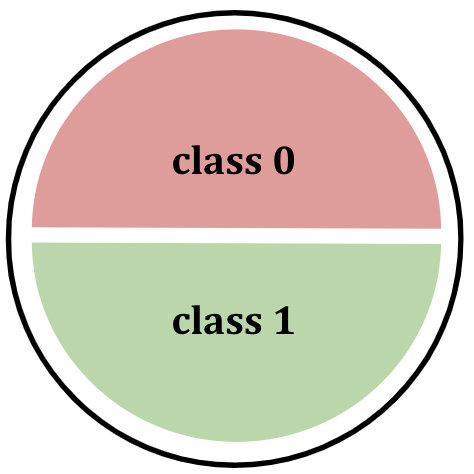} &\includegraphics[width=0.7\linewidth,align=c]{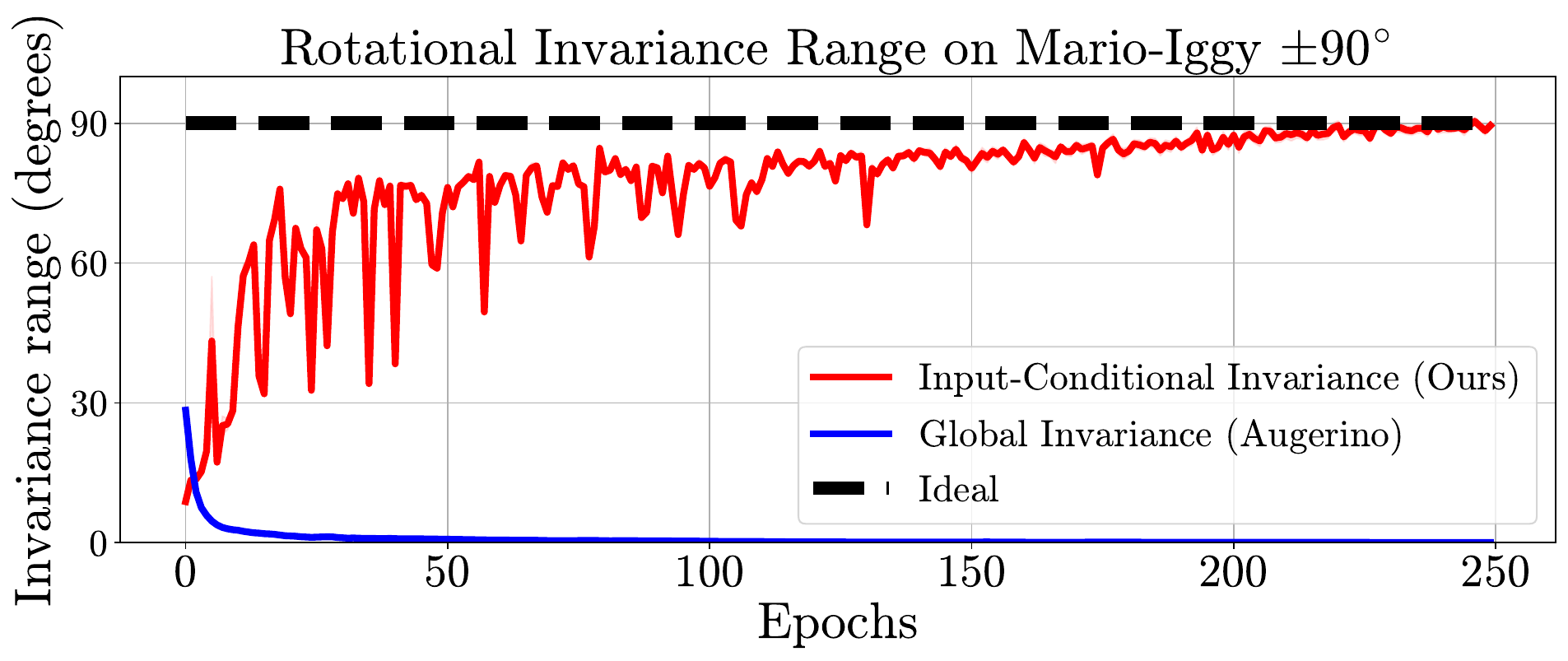} \\      
                \includegraphics[width=0.2\linewidth,align=c]{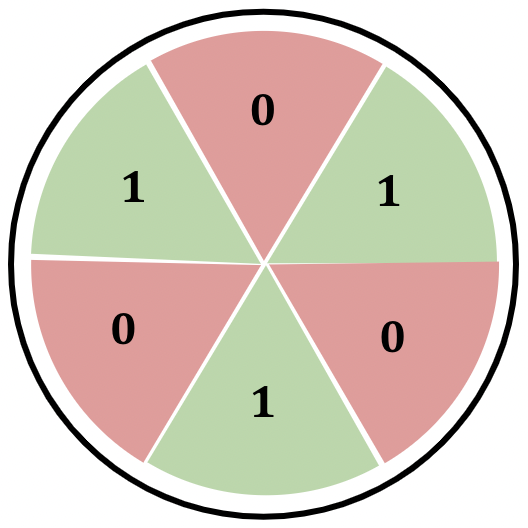} &\includegraphics[width=0.7\linewidth,align=c]{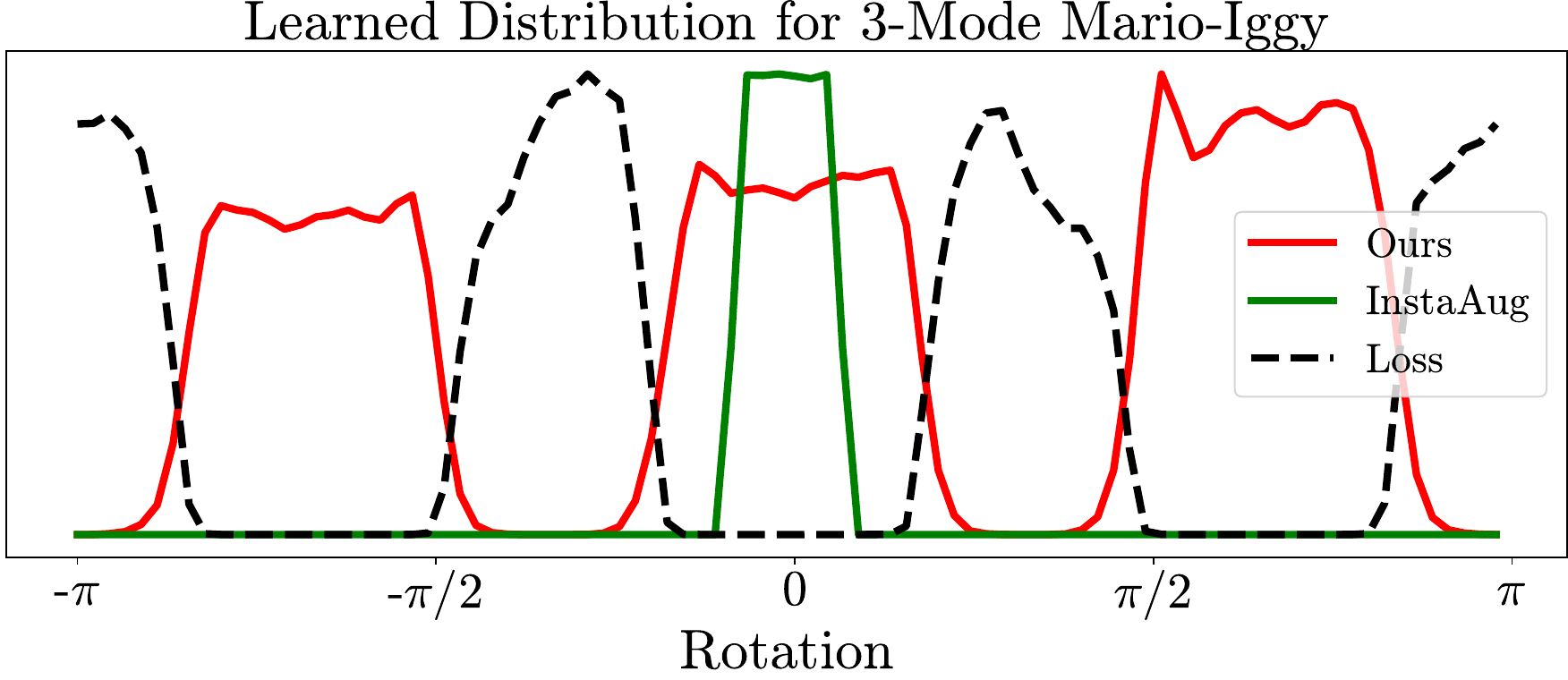} \\
        \end{tabular}

    \caption{Our normalizing flow model can represent input-dependent, multi-modal, and joint distributions over augmentation parameters. \textbf{(top)}  We illustrate three samples, each with a different set of correct augmentations. Augerino learns a range shared between all samples, so the learned range is too restrictive. InstaAug learns an instance-wise range but cannot handle a non-axis-aligned augmentation set (middle). In contrast, our model can adapt to the loss landscape and produce the largest possible set. 
    \textbf{(middle)} Augerino \cite{augerino} fails to learn augmentations in challenging settings. Learned rotation range for a version of Mario-Iggy with $\pm 90^{\circ}$ rotation range. The class boundaries touch each other, so some instances lie close to the boundary, and thus, global augmentation schemes like \cite{augerino, lila} are forced to learn a range of $0$. Our method learns the correct range. \textbf{(bottom)} InstaAug fails to capture the distribution for a multi-modal version of the Mario-Iggy dataset.} 
    \label{fig:distribution}
    \vspace{-6mm}
\end{figure}
}

\def\figCIFARLTres#1{
\begin{figure}[#1]
\centering
\includegraphics[width=0.9\linewidth]{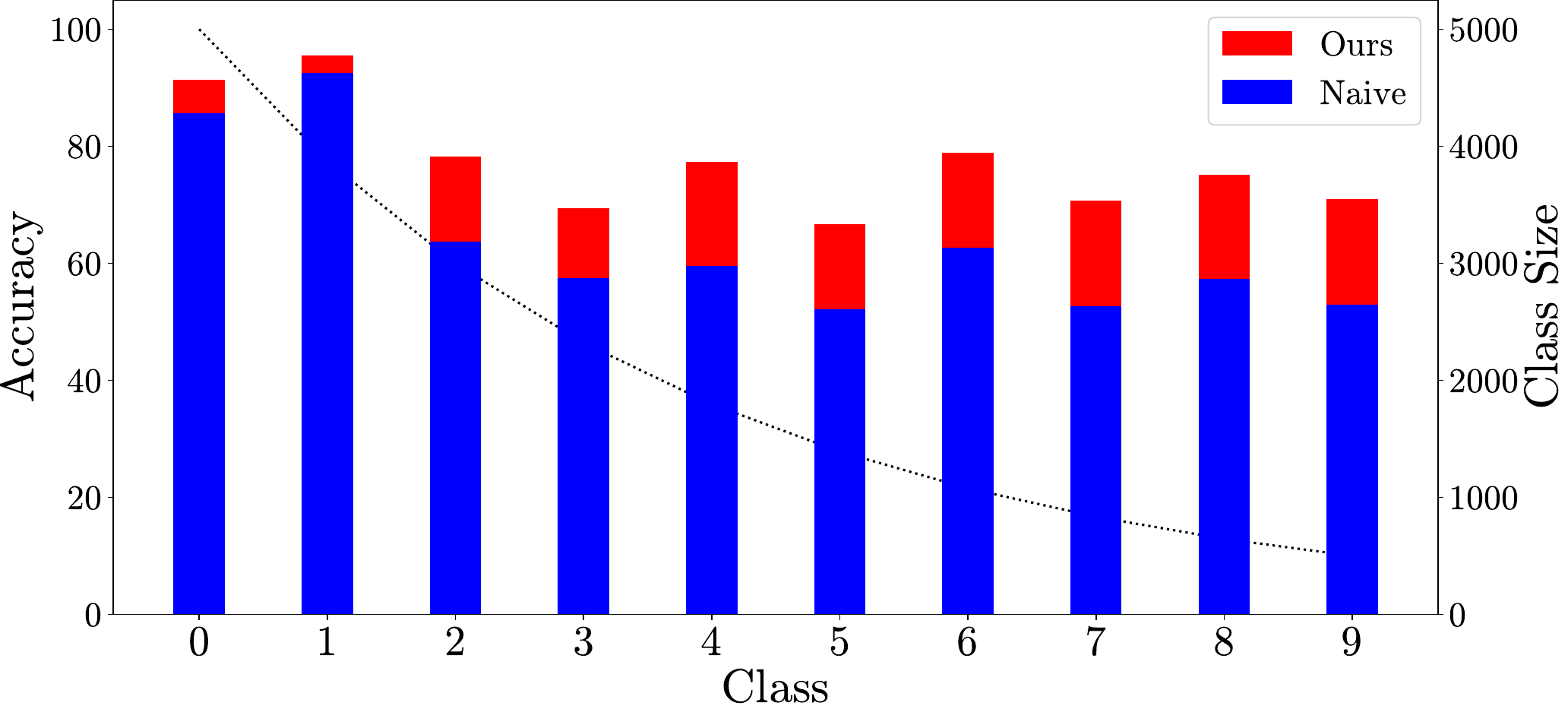}
\vspace{-3mm}
\caption{Our method delivers strong gains for imbalanced classification.  On CIFAR10-LT with 5000 to 500 instances per class from head to tail ({black curve}), our class-agnostic instance-wise transform distribution helps boost the classification accuracy by large margins (\textcolor{red}{red bars}) over the standard softmax baseline (\textcolor{blue}{blue bars}) especially for the tail classes. \label{fig:res}}
\vspace{-5mm}
\end{figure}
}

\usepackage[pagebackref=true,breaklinks=true,letterpaper=true,colorlinks,bookmarks=false]{hyperref}
\iccvfinalcopy %
\usepackage[capitalize,noabbrev]{cleveref}

\def\iccvPaperID{12600} %

\usepackage{graphicx}
\graphicspath{{images/}}

\ificcvfinal\fi

\begin{document}

\title{\mbox{Learning to Transform for Generalizable Instance-wise Invariance}}

\author{%
\setlength{\tabcolsep}{7pt}%
\begin{tabular}{@{}cccc@{}}
Utkarsh Singhal$^1$&
Carlos Esteves$^3$&
Ameesh Makadia$^3$&
Stella X. Yu$^{1,2}$ \\
\end{tabular}
\\
\setlength{\tabcolsep}{8mm}
\begin{tabular}{@{}ccc@{}}
$^1$ UC Berkeley&
$^2$ University of Michigan &
$^3$ Google Research \\
\end{tabular}
}
\maketitle
\ificcvfinal\thispagestyle{empty}\fi

\begin{abstract}

Computer vision research has long aimed to build systems that are robust to spatial transformations found in natural data. Traditionally, this is done using data augmentation or hard-coding invariances into the architecture. 
However, too much or too little invariance can hurt, and the correct amount is unknown a priori and dependent on the instance. Ideally, the appropriate invariance would be learned from data and inferred at test-time.

We treat invariance as a prediction problem. Given any image, we use a normalizing flow to predict a distribution over transformations and average the predictions over them. Since this distribution only depends on the instance, we can align instances before classifying them and generalize invariance across classes. The same distribution can also be used to adapt to out-of-distribution poses. This normalizing flow is trained end-to-end and can learn a much larger range of transformations than Augerino and InstaAug. When used as data augmentation, our method shows accuracy and robustness gains on CIFAR 10, CIFAR10-LT, and TinyImageNet. 

Project website is at: \url{https://sutkarsh.github.io/projects/learned_invariance}

\end{abstract}

\vspace{-5mm}
\section{Introduction}
\label{sec:intro}

One of the most impressive abilities of the human visual system is its robustness to geometric transformations. Objects in the visual world often undergo rotation, translation, etc., producing many variations in the observed image. Nonetheless, we classify them reliably and efficiently.

Any robust classifier must encode information about the expected geometric transformations, either explicitly (e.g., through architecture) or implicitly (e.g., invariant features). What would this knowledge look like for humans? 

Scientists have extensively investigated this question \cite{tarr1989mental}. We know that it generalizes to novel (but similar) categories, e.g., we can instantly recognize a new symbol from many poses after seeing it just once \cite{congealing}. For unfamiliar categories or poses, we can learn the invariance over time \cite{koriat1985mental}. Finally, while we quickly recognize objects in typical poses, we can also adapt to ``out-of-distribution'' poses with processes like mental rotation \cite{shepard1971mental}. These properties help us robustly handle novel categories and novel poses (\cref{fig:teaser}).

\figTeaser{t}
In contrast, modern classifiers based on deep learning are brittle \cite{spandan}. While these methods have achieved super-human accuracy on curated datasets like ImageNet \cite{deng2009imagenet}, they are unreliable in the real world \cite{shankar2021image}, showing poor generalization and even causing fatal outcomes in systems relying on computer vision \cite{uber}. Thus, robust classification has long been an aim of computer vision research \cite{spandan, geomDL}.  This paper asks:

\emph{Can we replicate this flexible, generalizable, and adaptive invariance in artificial neural networks?}

For some transformations (e.g., translation), the invariance can be hard-coded into the architecture. This insight has led to important approaches like Convolutional Neural Networks \cite{lecun1999object, fukushima1988neocognitron}. However, this approach imposes severe architecture restrictions and thus has limited applicability.

\figGraph{ht!}

An alternative approach to robustness is data augmentation \cite{grounding_inductive_biases}. Input data is transformed through a predefined set of transformations, and the neural network learns to perform the task reliably despite these transformations. Its success and wide applicability have made it ubiquitous in deep learning. However, data augmentation is unreliable. The learned invariance breaks under distribution shifts and doesn't transfer from head classes to tail classes in imbalanced classification~\cite{git}.

Both these approaches \emph{prescribe} the invariances while assuming a known set of transformations. However, the correct set of invariances is often unknown \textit{a priori}, and a mismatch can be harmful \cite{augerino, insta_aug, grounding_inductive_biases}. For instance, in fine-grained visual recognition, rotation invariance can help with flower categories but hurt animal recognition \cite{xiao2021what}.

A recent line of methods \cite{augerino, lila, insta_aug} aims to \emph{learn} the useful invariances. Augerino \cite{augerino} learns a range of transformations shared across the entire dataset, producing better generalizing models. However, these methods use a fixed range of transformations for all inputs, thus failing to be flexible. InstaAug \cite{insta_aug} learns an instance-specific augmentation range for each transformation, achieving higher accuracy on datasets such as TinyImageNet due to its flexibility. However, since InstaAug learns a range for each parameter separately, it cannot represent multi-modal or joint distributions (e.g., it cannot discover rotations from the set of all affine matrices). Additionally, these approaches don't explore generalization across classes and adaptation to unexpected poses (\cref{fig:teaser}).

We take inspiration from Learned-Miller~\etal \cite{congealing} and model the relationship between the observed image and its class as a graphical model (\cref{fig:graph,fig:graphicalmodel};). We also represent the instance-wise distribution of transformations using a normalizing flow and apply it to robust classification. Our experiments show that the properties like adaptability and generalizability emerge naturally in this framework.

\figCIFARLTres{tp}

\textbf{Contributions:} \textbf{(1)} We propose a normalizing flow model to learn the image-conditional transformation distribution. \textbf{(2)} Our model can represent multi-modal and joint distributions over transformations, being able to model more complex invariances, and \textbf{(3)} helps achieve higher test accuracy on datasets such as CIFAR10, CIFAR10-LongTail (\cref{fig:res}), and TinyImageNet. Finally, \textbf{(4)} combined with our graphical model, this model forms a flexible, generalizable, and adaptive form of invariance. It can be used to \textbf{(a)} align the dataset and discover prototypes like congealing \cite{congealing}, \textbf{(b)} adapt to unexpected poses like mental rotation \cite{koriat1985mental}, and \textbf{(c)} transfer invariance across classes like GAN-based methods \cite{git}.

\section{Related Work}

\textbf{Mental rotation in humans}: Shepard and Metzler~\cite{shepard1971mental} were among the first to measure the amount of time taken by humans to recognize a rotated object. They found that the response time increased linearly with rotation, suggesting a dynamic process like mental rotation for recognizing objects in unfamiliar poses.  Tarr and Pinker~\cite{tarr1989mental} further study mental rotation as a theory of invariant object recognition, contrasting it against invariant features and a multiple-view theory. Cooper and Shepard~\cite{cooper1973chronometric} found that revealing identity and orientation information beforehand helped the subjects make constant-time predictions. Hock and Tromley~\cite{hock1978upright} found that the recognition time is nearly constant for characters perceived as ``upright'' over a large range of rotations. However, outside that range (and for characters with narrow ``upright'' ranges), the recognition time follows the same linear relationship, indicating mental rotation is needed when the object is detected as ``not upright.'' Koriat and Norman~\cite{koriat1985mental} investigated mental rotation as a function of familiarity, finding that humans adapt to unfamiliar objects with practice, gaining robustness to small rotations around the upright pose. These works suggest a flexible, adaptive, and general form of robustness in the human vision.

\textbf{Invariance in Neural Networks}: Neural networks invariant to natural transformations have long been a central goal in deep learning research~\cite{geomDL}. Bouchacourt \etal~\cite{grounding_inductive_biases} and Madan \etal~\cite{Madan2022} studied the invariances present in modern models. One of the earliest successes includes architectures like Convolutional Neural Networks \cite{lecun1999object, fukushima1988neocognitron}, and more recently, applications such as medical image analysis~\cite{WinkelsC19,histopatho,GrahamER20}, cosmology~\cite{DielemanFK16,PerraudinDKS19}, and physics/chemistry~\cite{AndersonHK19,SchuttUG21,alphafold}. Kondor and Trivedi~\cite{kondor18icml} and Cohen \etal~\cite{cohen2019general} established a general theory of equivariant neural networks based on representation theory. 
Finzi \etal~\cite{lieconv} generalized convolutional neural networks to general Lie groups, and Residual Pathway Priors~\cite{finzi2021residual} combined equivariant and non-equivariant blocks through a residual connection. Transformed Risk Minimization \cite{TRM} is an alternative to ERM that minimizes the \emph{transformed} population risk, providing strong PAC-Bayes bounds on generalization.

The dominant way to add invariance into neural networks is data augmentation. Dao~\etal~\cite{Dao2019AKT} shows that to a first-order approximation, data augmentation is equivalent to averaging features over transformations. Bouchacourt \etal~\cite{grounding_inductive_biases} found data augmentation to be crucial for invariance in many modern architectures. Zhou \etal~\cite{git} demonstrated a key failing of data augmentation in imbalanced classification and used a GAN to generate a broad set of variants for every instance. Our method is complementary to theirs and can be combined in future work. We also note that the experiments in this paper only use affine image transformations and yet achieve comparable accuracy to theirs on CIFAR10LT. Congealing~\cite{congealing} aligns all the images in a class, simultaneously producing a prototype and inferring the relative pose of each example. The aligned dataset can be used for robust recognition, and the learned pose distribution can be used for new classes. However, this method assumes the transformation distribution is class-wise, whereas we model it for every instance. Learned canonicalization \cite{kaba2022equivariance} learns an energy function that is minimized at test time to align the input to a canonical orientation. Spatial Transformer Networks \cite{jaderberg15spatialtransformer} predict a transformation from the input image in an attempt to rectify it and improve classification accuracy. However, STNs cannot represent a distribution of transformations. Probabilistic Spatial Transformer Networks \cite{pstn} model the conditional distribution using a Gaussian distribution with mean and variance predicted by a neural network. In contrast, we use a normalizing flow model. We also study generalizability and adaptation to unexpected poses.

\figDistribution{t}

\textbf{Augerino}: \cite{augerino} aims to learn the ideal range of invariances for any given dataset. It uses the reparametrization trick and learns the range of uniform distribution over each transformation parameter separately (e.g., range of translations, rotations, etc.). This ability allows Augerino to learn the useful range of augmentations (and thus invariances) directly and produce more robust models with higher generalization. However, Augerino is sensitive to the regularization amount and the parametrization of the augmentation range (\cref{tab:tin}). LILA~\cite{lila} tackles this problem using marginal likelihood methods. However, for both Augerino and LILA, the resulting invariance is shared among all classes, even though different classes (such as $0$ and $6$ in a digit classification setting) may have entirely different ideal augmentation distributions. \cref{fig:distribution} illustrates how these limitations lead Augerino to learn an overly restricted augmentation range.

\textbf{InstaAug:} \cite{insta_aug} fixes the inflexibility of Augerino by predicting the augmentation ranges for every instance and provides a theoretical argument connecting it to generalization error. This allows for larger effective ranges and, thus, impressive generalization gains in image classification and contrastive learning settings. However, while InstaAug is instance-wise, it models the range of each parameter separately (the \emph{mean-field} assumption). Thus, it cannot represent multi-modal or joint distributions (\cref{fig:distribution}). Like Augerino, the set of learnable transformations is greatly limited, especially for complex augmentation classes like image cropping \cite{insta_aug}, necessitating tricks like selecting among a pre-defined set of crops. Furthermore, InstaAug is sensitive to parametrization (see \cref{fig:orth,tab:tin}).

\section{Methods}
We begin by describing our probabilistic model. We derive its inference equation and training loss and compare it to existing methods. We then construct a normalizing flow model to represent the conditional transform distribution. We also derive an analytical expression for the model's approximate invariance. Finally, we describe the mean-shift algorithm for adapting to out-of-distribution poses.

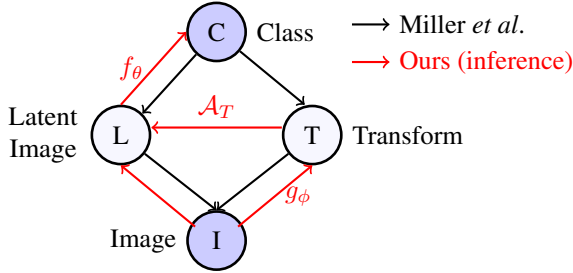
\begin{figure}[ht!]
    \vspace{-2mm}
    \centering
            \begin{tikzpicture}[node distance=1.8cm, thick]
                \tikzstyle{observed} = [circle, draw, fill=blue!20, line width=1pt, minimum size=2.2em, text=black];
                \tikzstyle{latent} = [circle, draw, fill=blue!3, line width=1pt, minimum size=2.2em, text=black];
                
                \node[observed, draw, fill=blue!20, label=right:{Class}] (C) {C};
                \node[latent, below left = 0.8cm and 0.7cm of C, label={[text width=1cm, align=center]left:Latent\\ Image}] (L) {L};
                \node[latent, below right = 0.8cm and 0.7cm of C, label=right:{Transform}] (T) {T};
                \node[observed, below=1cm of $(L)!0.5!(T)$, label=left:{Image}] (I) {I};
                
                \draw[->] (C) -- ($(L.north east)$);
                \draw[->] (C) -- ($(T.north) - (0.1, 0)$);
                \draw[->] (L) -- ($(I.north)$);
                \draw[->] (T) -- ($(I.north)$);
                
                \draw[->, red] ($(L.north)$) -- node[midway, left, color=red] {$f_{\theta}$} ($(C.west)$);
                \draw[->, red] ($(T.west) + (0, 0.1)$) -- node[midway, above, color=red] {$\mathcal{A}_T$} ($(L.east) + (0, 0.1)$);
                \draw[->, red] ($(I.north west)+ (0, -0.1)$) -- ($(L.south)$);
                \draw[->, red] ($(I.north east)+ (0, -0.1)$) -- node[midway, right, color=red] {$g_{\phi}$} ($(T.south)$);
                
                \begin{scope}[xshift=1.8cm,yshift=0.1cm]
                    \draw[->,black] (0,0) -- ++(0.5,0) node[right, black]{Miller~\etal};
                    \draw[->,red] (0,-0.5) -- ++(0.5,0) node[right, red]{Ours (inference)};
                \end{scope}
            \end{tikzpicture}        
    \caption{Our graphical model inspired by Miller~\etal\cite{congealing}. Shaded nodes represent variables observed in data ($C, I$). In contrast to Miller~\etal, we only model the inference process and assume that $T$ is instance-wise, not classwise. Our flow model $g_\phi$ predicts image-conditional transform, and the classifier $f_\theta$ classifies the resulting image $L$.}
    \vspace{-3mm}
    \label{fig:graphicalmodel}
\end{figure}

\textbf{Graphical model}: We follow the model described in \cref{fig:graphicalmodel}. Here, $C$ refers to the class, $I$ refers to the observed image, $L$ refers to the latent image (equivalent to the prototype in \cite{congealing}), and $T$ refers to the unobserved transformation parameters connecting the latent image and the observed image. The latent image is produced by passing the pair $(I, T)$ through a differentiable augmenter $\mathcal{A}$, which applies the transform to the observed image, i.e., $L = \mathcal{A}_T(I)$.

One notable difference to Miller~\etal\cite{congealing} is that our distribution is instance-wise (similar to \cite{insta_aug}), not class-wise. This allows for a more general conditional distribution model.

Given the values $C,L,T,I$, the model defines a joint probability distribution $P(C, L, T, I)$: \begin{align}
P(C, L, T, I) = P(C\lvert L)P(L\lvert T,I)P(T\lvert I)P(I)
\end{align}
and the conditional class probability $P(C|I)$ as:
\begin{align}
P(C | I) = \int_{L,T}P(T \lvert I)P(L\lvert T,I)P(C\lvert L) dLdT
\end{align}

Since $L = \mathcal{A}_T(I)$, this can be further simplified to: \begin{align}
P(C | I) &= \int_{T}P(T \lvert I)P(C\lvert L=\mathcal{A}_T(I)) dT \\
&= \mathbb{E}_{T \sim P(T \lvert I)} \big[ P(C\lvert L=\mathcal{A}_T(I)) \big]
\end{align}
Thus, the predicted class probability is averaged over transformations sampled from the conditional transform distribution $P(T \lvert I)$. This is analogous to the idea of ``test-time augmentations'' used in image classification literature.  Augerino assumes that the transformation $T$ is independent of $I$. InstaAug models $T$ as a uniform distribution conditioned on $I$.  PSTN \cite{pstn} arrives at the same expression and uses a Gaussian distribution. All these frameworks can be viewed as different approximations in this formulation. However, we also analyze the invariance properties of this formulation and applications of $P(T|I)$.

\textbf{Neural network approximation}: We approximate each of the key distributions $P(C\lvert L)$ and $P(T\lvert I)$ with neural networks. Our $f_{\theta}(C;L)$ is a simple classifier, and $g_{\phi}(T; I)$ is a normalizing flow model \cite{nflow} which takes in the image $I$: \vspace{-5mm} \begin{align} f_{\theta}(C;L)  \approx P(C \lvert L), \quad g_{\phi}(T;I)  \approx P(T \lvert I) \end{align}
Since $L = \mathcal{A}_T(I)$, we use $f_{\theta}(C;L)$, $f_{\theta}(C;T,I)$  and $f_{\theta}(C;\mathcal{A}_T(I))$ interchangeably.

\textbf{Inference:} The expression for $P(C \lvert I)$ then becomes:
\vspace{-5mm} \begin{align} p_{\theta, \phi}(C|I) &= \int_T g_{\phi}(T; I) f_{\theta}(C; \mathcal{A}_T(I)) dT \\
&= \mathbb{E}_{T \sim g_{\phi}(T; I)} \big[ f_{\theta}(C;\mathcal{A}_T(I)) \big] \end{align}
This equation describes the act of sampling transformations from the normalizing flow model and averaging the classifier predictions over the sampled transformations.

\textbf{Classifier loss}: During training, we observe $(I, C)$ pairs. We train the classifier $f_{\theta}$ by maximizing a lower bound to the average $\log p_{\theta, \phi}(C|I)$. It is common to use Jensen's inequality to make this tractable:
\begin{align}
    \log p_{\theta, \phi}(C|I) 
    \geq  \mathbb{E}_{T \sim g_{\phi}(T; I)} \big[\log f_{\theta}(C; \mathcal{A}_T(I))\big]
\end{align}
and maximize the resulting lower bound instead. This further reduces to the loss function $\mathcal{L}_{\operatorname{classifier}}$:
\begin{align}
    \mathcal{L}_{\operatorname{classifier}}  = \mathbb{E}_{T \sim g_{\phi}(T; I)} \big[-\log f_{\theta}(C; \mathcal{A}_T(I))\big]
\end{align}
which is simply the cross-entropy loss averaged over sampled augmentations.
\begin{figure}
    \centering
     \includegraphics[width=0.95\linewidth]{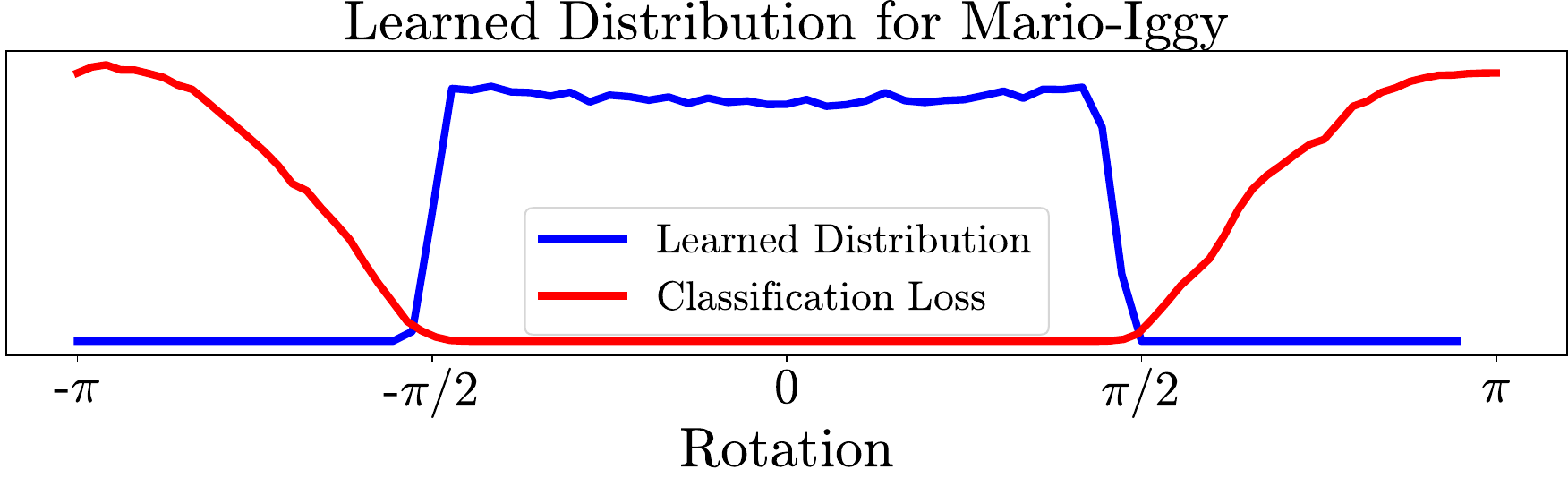}  \vspace{-3mm}
    \caption{The ideal learned distribution maximizes the range while minimizing the overall classification loss}
    \label{fig:lossdist}
    \vspace{{-4mm}}
\end{figure}

\textbf{Augmenter loss}: Intuitively, we would like the transform distribution $g_\phi$ to have a large diversity of augmentations and minimal classification loss (see \cref{fig:lossdist}). However, in practice, minimizing the classification loss leads to $g_{\phi}$ collapsing to a single peak ($0$-variance distribution) as the model overfits to the training data (as observed in Augerino \cite{augerino} without regularization). 

Since our normalizing flow model already produces log probability for each generated sample, \emph{entropy regularization} is a natural match for our method. We penalize the average $\log g_{\phi}$ for sampled transformations: \begin{align}
 \mathcal{L}_{\operatorname{augmenter}} = \mathcal{L}_{\operatorname{classifier}} + \alpha \mathbb{E}_{T \sim g_{\phi}} \big[\log g_{\phi}(T;I)\big]
\end{align}

This regularization is a generalization of the one used by Augerino, since for uniform distributions, $\log p \propto -\log(\text{width})$. InstaAug derives a similar expression as a Lagrange relaxation of entropy constraints and applies it to simple distributions like uniform and categorical.

We apply it to normalizing flows, which can model more general distributions, and our graphical model helps us understand this loss and connect it to variational inference.

\textbf{Understanding entropy regularization}: Here, we analyze the form of the distribution learned through entropy regularization. Consider the following loss: \vspace{-2mm}
\begin{align}
      \mathcal{L}_{\operatorname{augmenter}} [g_\phi] = \mathcal{L}_{\operatorname{classifier}} [g_\phi] - \alpha \mathbb{H}[g_{\phi}]
\end{align}
where $\alpha \in \mathbb{R}^+$ is a regularization constant and $\mathbb{H}[g_{\phi}]$ is the entropy of the distribution $g_{\phi}$. This expression reduces to: 
$$ = \mathbb{E}_{T \sim  g_{\phi}(T;I)} \big[  \alpha \log g_\phi(T;I) - \log f_{\theta}(C;\mathcal{A}_T(I)) \big]   $$
We rescale this loss by $\lambda = \frac{1}{\alpha}$ to simplify:
$$ \equiv \mathbb{E}_{T \sim  g_{\phi}(T;I)} \big[ \log g_\phi(T;I) - \lambda \log f_{\theta}(C;\mathcal{A}_T(I)) \big]  $$
Note that this loss is equivalent to a KL-divergence between $g_\phi$ and a special target distribution $\tilde{p}^{\lambda}_{\theta}(T \lvert C,I)$:
$$ \mathcal{L}_{\operatorname{augmenter}} [g_\phi]  = \operatorname{KL}\big[ g_\phi(T;I) ||\ \tilde{p}^{\lambda}_{\theta}(T \lvert C,I) \big]  $$
where the target distribution $\tilde{p}^{\lambda}_{\theta}(T \lvert C,I)$ is defined as:
\vspace{-1mm}
\begin{align*}
  \tilde{p}^{\lambda}_{\theta}(T \lvert C,I) &= \frac{1}{Z(\lambda)} f_{\theta}(C; T, I)^{\lambda}
\end{align*}
where $Z\in \mathbb{R}^+$ is a normalization constant and $\lambda \in \mathbb{R}^+$ is a temperature constant. This distribution is formed by computing $p_{\theta,\phi}(C \lvert T, I)^{\lambda}$ over transforms $T$ and normalizing them. Thus, it assigns a higher probability to the transformations with lower classification loss. $\lambda$ here is analogous to the temperature parameter in softmax, and large values of $\lambda$ make the distribution highly peaked. In contrast, small values suppress peaks and make the distribution less ill-behaved as a target. $\lambda \to 0$ corresponds to a uniform distribution, whereas $\lambda \to \infty$ collapses the distribution to the single transformation that minimizes the classification loss.

We also note that when $\lambda=1$, the target distribution $\frac{1}{Z} p_{\theta}(C \lvert T, I)$ is exactly the posterior $p_{\theta}(T \lvert C, I)$, assuming a uniform prior for the unknown $p_{\theta}(T|I)$. Different choices of this prior lead to other loss functions, like a Gaussian prior penalizing the transformation norm. However, we stick to the uniform prior for simplicity. %

\textbf{Representing the conditional distribution}: Our approach uses parametrized differentiable augmentations similar to Augerino. However, instead of learning the global range of transformations, we predict a distribution over the transformations conditioned on the input image. We use an input-conditional normalizing flow model \cite{nflow}.

A normalizing flow model starts with a simple pre-defined probability distribution $p_0$, e.g., Normal distribution. For a sample $z_0 \sim p_0$, it successively applies transformations $f_1,f_2, \ldots, f_K$, producing a more complicated distribution by the end. The log probability density of the final sample is given by $\log p(z_k)$ = $\log p_0(z_0) - \log | \det \frac{dz_k}{dz_0} |$, and the architecture is designed to allow efficient sampling and computation of $\log p$. We use the samples to augment the input (\cref{fig:graph}) and $\log p$ term in the loss. Our model is based on RealNVP \cite{dinh2017density}, using a mixture of Gaussians as the base $p_0$.

Given any input image $I$, we use a convolutional feature extractor to extract an embedding vector $e$. This embedding vector is then projected down to a scale and bias used by each layer of the normalizing flow and the base distribution. This normalizing flow model outputs samples $s$ from the augmentation distribution and their corresponding log-probabilities $\log p(s)$. These samples are passed to the differentiable augmentation, which transforms the input image to be processed by the model (\cref{fig:graph}) using PyTorch's $\textit{grid\_sample}$. While we use affine image transformations for our experiments, our method generalizes to any differentiable transformation.

\textbf{Approximate invariance}: Here, we formalize the notion of approximate invariance and connect it to our classifier and flow model. Intuitively, the approximate invariance in our method comes from both the augmenter and the classifier. Their contributions can be divided into (1) the classifier's inherent insensitivity to transformations, (2) the width of the transform distribution being used for averaging, and (3) the canonicalization effect of the transform distribution. Each of these properties corresponds to a different theory of object recognition explained by Tarr and Pinker~\cite{tarr1989mental} and connected to deep neural networks by Kaba~\etal\cite{kaba2022equivariance}. We formalize this intuitive argument as follows: Given an input image $I$, our model's output is the classifier prediction averaged over $g_{\phi}(T;I)$, i.e. $p_{\theta, \phi}(C|I) =  \mathbb{E}_{T \sim g_{\phi}(T; I)} \big[ f_{\theta}(C;\mathcal{A}_T(I)) \big]$ (see equation 9). Let a new image $I'$ be formed by transforming the original image by a transformation $\Delta T$, i.e. $ I' = \mathcal{A}_{\Delta T}(I)$. Then:
$$ p_{\theta, \phi}(C|I') = \int_T g_{\phi}(T; \mathcal{A}_{\Delta T}(I)) f_{\theta}(C;\mathcal{A}_{T+\Delta T}(I)) dT $$
$$  =  \int_T g_{\phi}(T-\Delta T; I') f_{\theta}(C;\mathcal{A}_{T}(I)) dT $$
Where the last step substitutes $T$ for $T + \Delta T$. Then, the change in prediction,  denoted as $\operatorname{err}(C; I, I')$, is:
$$\operatorname{err}(C; I, I') =  | p_{\theta, \phi}(C|I) - p_{\theta, \phi}(C|I') | $$
$$ = \Bigl| \int_T \big[ g_{\phi}(T-\Delta T; I')-g_{\phi}(T;I) \big] f_{\theta}(C;\mathcal{A}_{T}(I)) dT \Bigr| $$
Next, we derive bounds on this quantity based on $g_\phi$ and $f_\theta$.  Let $S = \operatorname{supp}(g_\phi(.;I)) \cup \operatorname{supp}(g_\phi(.;I'))$ is the support set of the transform distributions, i.e. all the samples for $I$ and $I'$ are inside $S$.  We can thus limit the integration to $S$:
$$ = \Bigl| \int_{T \in S} \big[ g_{\phi}(T-\Delta T; I')-g_{\phi}(T;I) \big] f_{\theta}(C;\mathcal{A}_{T}(I)) dT \Bigr| $$
Let's now quantify the behavior of $f_\theta$ on $S$. Let $M$ be the maximum and $m$ be the minimum of $f_\theta$ on this set, i.e. 
$$ M = \max_{t \in S} f_{\theta}(C;\mathcal{A}_{T}(I)), \quad m = \min_{t \in S} f_{\theta}(C;\mathcal{A}_{T}(I)),$$
Note that the first term $g_{\phi}(T-\Delta T; I')-g_{\phi}(T;I)$ is the difference of two probability density functions and so integrates to $0$. Thus, if we add a constant value to $f_\theta$, it doesn't change the whole integral. Subtracting $m$, we get:
\begin{align*}
\Bigl| \int\displaylimits_{T \in S} \big[ g_{\phi}(T-\Delta T; I')-g_{\phi}(T;I) \big] (f_{\theta}(C;T,I)-m) dT \Bigr|
\end{align*}
Using $\lvert \int f(x) dx \rvert \leq \int \lvert f(x) \rvert dx$ and $\lvert x y \rvert = \lvert x \rvert \lvert y \rvert$ we have: \begin{align*}
\centering
\leq &\int\displaylimits_{T \in S}  \Bigl| g_{\phi}(T-\Delta T; I')-g_{\phi}(T;I) \Bigr|\Bigl| f_{\theta}(C;T,I)-m \Bigr| dT \\
\leq & (M-m) \int\displaylimits_{T \in S}  \Bigl| g_{\phi}(T-\Delta T; I')-g_{\phi}(T;I) \Bigr| dT \\
= & 2(M-m) \operatorname{TV}[g_{\phi}(T-\Delta T; I')\| \ g_{\phi}(T;I)]
\end{align*}
where $\operatorname{TV}$ refers to the Total Variation Distance defined as $\operatorname{TV[p\| q]}=\frac{1}{2}\int |p(x)-q(x)|dx$. In summary:
$$\operatorname{err}(C; I, I') \leq 2(M-m) \operatorname{TV}[g_{\phi}(T-\Delta T; I')\| g_{\phi}(T;I)]$$
Thus, the prediction change ($\operatorname{err}(C; I, I')$) is upper bounded by two factors: \textbf{(1)} $M-m$, which measures how much the classifier predictions change over the relevant range, and \textbf{(2)} the total variation distance between the original transform distribution $g_{\phi}(T;I)$ and the new version  $g_{\phi}(T-\Delta T; I')$. This result explains how the method achieves approximate invariance. If the classifier features are invariant to the input transformations, we get $M-m \approx 0$, and thus $error \approx 0$. The same is true if the transform distribution is approximately equivariant, i.e. $g_{\phi}(T-\Delta T; I') \approx g_{\phi}(T;I)$.

\textbf{Mean-shift for handling out-of-distribution poses}: While the conditional transformation distribution $g_{\phi}(T;I)$ can adjust to in-distribution pose variation, this approach does not work for out-of-distribution poses (see \cref{fig:test_time_robust}). We use a modified version of the well-known \emph{mean-shift algorithm}. Instead of sampling points from a dataset and weighting them with a kernel, we directly use $g_{\phi}$ samples.

The core idea is to push the image closer to a local mode where our models may work better. We start with image $I_0$ and the transform parameter $T_0 = 0$. Then, at every step: $$ T_k :=  T_{k-1} + \gamma \mathbb{E}_{T \sim g_{\phi}(T;I_{k-1})}[T], \quad I_k := \mathcal{A}_{T_k}(I_{0}) $$
where $\gamma \in \mathbb{R}^+$ is the step size. In summary, the algorithm repeatedly samples from the conditional distribution, computes the mean, and accumulates the result into $T$.

Since our method learns an input-conditional probability distribution, the mean of the augmentation transformation $\mathbb{E}_{T \sim g_{\phi}(T;I)}[T]$ for any given image is an estimate of the difference between the local mode and the current transform $T$. Thus, each step moves the image closer to the local mode, which is the fixed point for this process.

\section{Experiments}
We benchmark accuracy on datasets such as CIFAR10 and TinyImageNet, and plot the learned transformation distribution for toy examples on Mario-Iggy \cite{augerino} and MNIST. Finally, we test applications of the learned distribution. The code and scripts to reproduce all the results can be found at \url{https://github.com/sutkarsh/flow_inv/}

\tabLILA{ht!}

\textbf{CIFAR10}: We benchmark our method against Augerino and LILA \cite{lila} on learning affine image transformations for CIFAR10 classification. We use the models and libraries provided by \cite{lila}. We use a RealNVP flow \cite{dinh2017density} with permutation mixing, $12$ affine coupling layers, and a 2-layer MLP of width $32$ for each layer. We turn the input into an embedding using a 5-layer CNN and append this embedding to each layer's MLP input as well as project it to the parameters of the base distribution, which is a mixture of Gaussians. We also add a $\operatorname{tanh}$ at the end of the flow to ensure the produced distribution stays within bounds. Please see the supplementary material for more details. Using a modified ResNet18~\cite{HeZRS16}, and train our model for $200$ epochs. We report the accuracy in \cref{tab:lila}. Our method is able to achieve a $7.8\%$ test accuracy gain compared to Augerino and $2.6\%$ against LILA. We note that our method is still based on maximum likelihood; thus, LILA's marginal likelihood method is complementary to ours. These methods may be combined for even higher accuracy in future work. We also report the accuracies for MNIST and FashionMNIST.

\textbf{Imbalanced CIFAR-10 Classification}: Imbalanced classification is a challenging setting for invariance learning. As shown by \cite{git}, invariances learned through data augmentation do not transfer from head classes to tail classes. This is especially harmful since the tail classes, due to a small number of examples, benefit the most from the invariance. CIFAR10-LT is an imbalanced version of CIFAR10 where the smallest class is $10\text{x}$ smaller than the largest. Here, we outperform Augerino by $14.5\%$ and LILA by $1.7\%$.

\textbf{Augerino 13-layer CIFAR10}: We also evaluate our method on Augerino’s 13-layer network, re-using the same hyperparameters as the LILA experiments \cref{tab:lila}. Our method achieves $94.3\%$ test accuracy ($0.5\%$ gain).
\begin{table}[hb!]
    \vspace{-2mm}
    \centering
    \scalebox{0.99}{
    \begin{tabular}{ccccc}
    \toprule
       & No Aug.  & Fast AA &  Augerino & Ours \\
    \midrule
      Acc & 90.6 & 92.65 & 93.8 & \textbf{94.3} \small{$\pm$ 0.08} \ \\
    \bottomrule
    \end{tabular}}
    \vspace{-2mm}
    \caption{Test accuracies for Augerino's 13-layer model. Baseline numbers quoted from \cite{augerino}.}
    \vspace{-3mm}
\end{table}

\textbf{TinyImageNet Classification}: We evaluate our method against InstaAug on the TinyImageNet dataset. This 64x64 dataset contains 200 classes. The goal of this task is to learn cropping augmentations. A crop can be parametrized with four parameters: $(\text{center}_x, \text{center}_y, \text{width}, \text{height})$, so we represent it with a 4-dimensional distribution. Please see the supplementary material for more details.

Cropping is a challenging augmentation to learn since the crop location and size are correlated. InstaAug's mean-field representation cannot represent this, so achieves low accuracy without the location-related parameterization (LRP). LRP consists of $321$ pre-defined crops and predicts the probability of each crop. This approach does not scale to high dimensional distributions (e.g. specifying more transformations). In contrast, our method can achieve high accuracy without LRP, beating InstaAug by nearly $11\%$ (\cref{tab:tin}).

\begin{table}[h]
    \centering
    \begin{tabular}{l|r|r}
        \toprule
        \textbf{Method} &\textbf{Acc (\%)} &\textbf{+LRP(\%)} \\
        \midrule
        Baseline & 55.1 & --- \\
        Random Crop & \underline{64.5} & --- \\
        \midrule
        Augerino & 55.0 & --- \\
        InstaAug & 54.4 & \underline{66.0} \\
        Ours & \textbf{65.4} & \underline{66.0} \\
        \bottomrule
    \end{tabular}
\caption{\label{tab:tin} TinyIN classification accuracy on PreActResNet used by InstaAug, with and without location-relation parameterization. InstaAug is limited by its mean-field representation, performing poorly without LRP. In contrast, our method performs well regardless of parametrization.}
\vspace{-3mm}
\end{table}

\textbf{Learned invariance visualization} Mario-Iggy is a toy dataset from \cite{augerino} consisting of rotated versions of two images. Upright/upside-down images are classified as different classes, and each sample lies within $\pm 45^{\circ}$ of its class prototype. Since the range of rotations can be altered, this dataset is useful for studying learned invariance.  We consider two variations: \textbf{$\pm 90^{\circ}$ range}, and \textbf{multi-modal with $3$ modes}. 

The ideal augmentation distribution for Mario-Iggy dataset is $\pm 90^{\circ}$ around the class prototype. As the input image rotates, the augmentation distribution shifts such that the resulting augmented image distribution is constant. Our model trained on Mario-Iggy can reliably learn an invariant augmentation distribution (\cref{fig:distribution}). In the challenging multimodal distribution setting, our model can represent the three modes, whereas InstaAug fails.

\figorth{t!}
\figMNIST{t!}

\textbf{Representing joint distributions}: We test the ability of our normalizing flow to represent joint distributions by intentionally sampling from a larger set of transformations and letting the model learn the useful subset. Specifically, we start from the Lie algebra parametrization of affine transforms (used by Augerino). For rotation by $r$ radians, the transformation matrix is:
\vspace{-5mm}
\begin{align}
    T_{\operatorname{Augerino}}(r) = \exp \left( \begin{bmatrix}
 0 & r & 0\\
-r & 0 & 0\\
0 & 0 & 1
\end{bmatrix} \right)
\end{align}
For this experiment, we generalize this formulation as:
\vspace{-2mm}
\begin{align}
    T_{\operatorname{Decoupled}}(a,b,c,d,e,f) = \exp \left( \begin{bmatrix}
 a & b & c\\
d & e & f\\
0 & 0 & 1
\end{bmatrix} \right)
\end{align}
This matrix represents a rotation if $b=-d$. Since the Mario-Iggy dataset only contains rotations, the goal is to produce samples such that $b=-d$. Samples that do not follow this constraint will be out-of-distribution. \cref{fig:orth} shows that, unlike our model, InstaAug \cite{insta_aug} fails to learn rotation transforms for Mario-Iggy, even though skewed samples incur a higher loss. This is due to InstaAug's mean-field model, which predicts the range for each parameter separately, thus preventing it from following the $b=-d$ constraint. In contrast, our model learns to represent this joint distribution. We further test our model's ability to learn the rotation constraint on all $6$ affine parameters. \cref{fig:orth} also shows the deviation of sampled transformations from a true rotation matrix. Our learned distribution is concentrated close to the rotation transformations, showing that our method can start from a large group of transformations and learn to constrain it to only what is useful for the dataset and task.

\figInvarianceTransferClass{t!}

\textbf{Learning selective invariance for MNIST}: We test our model's selective invariance ability on the MNIST dataset (specifically $0$,$1$,$5$,$6$,$9$) and visualize the augmentation range for a few examples as well as class averages (see \cref{fig:mnist}). For digits $0$, $1$, and $5$, which can be recognized from any rotation, the learned rotation range corresponds to the entire $360^{\circ}$, whereas for $6$ and $9$, which may be confused with each other, the range is only $180^{\circ}$. In contrast, augerino learns a constant range. We find the same trend at the class level.

\figTTRobust{t!} 
\textbf{Generalizing invariance across classes}: Zhou~\etal~\cite{git} shows that invariances learned from head classes fail to transfer to tail classes. This is a major drawback of traditional data augmentation. We test generalization across classes by plotting the same metric as \cite{git} (expected KL divergence) across a range of rotations for CIFAR10-LT and RotMNIST-LT classifiers. Since RotMNIST-LT is a rotationally invariant dataset, we rotate all the images randomly in the $\pm 180^\circ$ range, whereas for CIFAR10-LT we use a $\pm10^\circ$ range. Our model achieves significantly lower eKLD, especially for tail classes (\cref{fig:transfer_class}), indicating higher robustness.

\textbf{Aligning image datasets like in Congealing \cite{congealing}}: We apply the mean-shift algorithm using the augmentation distribution trained on the Mario-Iggy ($45^{\circ}$) dataset. The Mario-Iggy dataset contains rotated versions of the Mario image with one unknown prototype, making it ideal for this test.

For each image, we apply the mean-shift algorithm. Each step moves the image closer to the local mode. We apply this procedure for $50$ iterations for every image separately. This process results in all the images in a small neighborhood agglomerating to the local prototype (\cref{fig:alignment}). \footnote{This draft omits a subfigure originally included in the ICCV version. Please see the errata in the appendix for details.}

We also tested this approach on MNIST, an out-of-distribution dataset for the mario-iggy model, and added $\pm45^{\circ}$ rotations for an additional challenge. Surprisingly, the method still aligns images and discovers prototypes (\cref{fig:alignment}) despite not being trained on any MNIST images.

\textbf{Summary}: We propose normalizing flows to learn the instance-wise distribution of transformations. It helps us make robust  classifiers, perform test-time alignment, discover prototypes, transfer invariance, and achieve higher test accuracy. These results highlight the potential of flexible, adaptive, and general invariance in computer vision.

\textbf{Acknowledgements}: This work was supported, in part, by the BAIR/Google fund.

{\small
\bibliographystyle{unsrt}
\bibliography{references}
}

\clearpage
\newpage
\appendix

\section{Supplementary Material}
We \textbf{(a)} present a class-wise accuracy analysis for CIFAR10-LT results, \textbf{(b)} present further analysis of the alignment experiment, discussing failure modes, \textbf{(c)} share experimental details, and \textbf{(d)} provide an errata.

\subsection{CIFAR-10LT class-wise accuracy}
\begin{figure}[h]
    \centering
    \includegraphics[width=\linewidth]{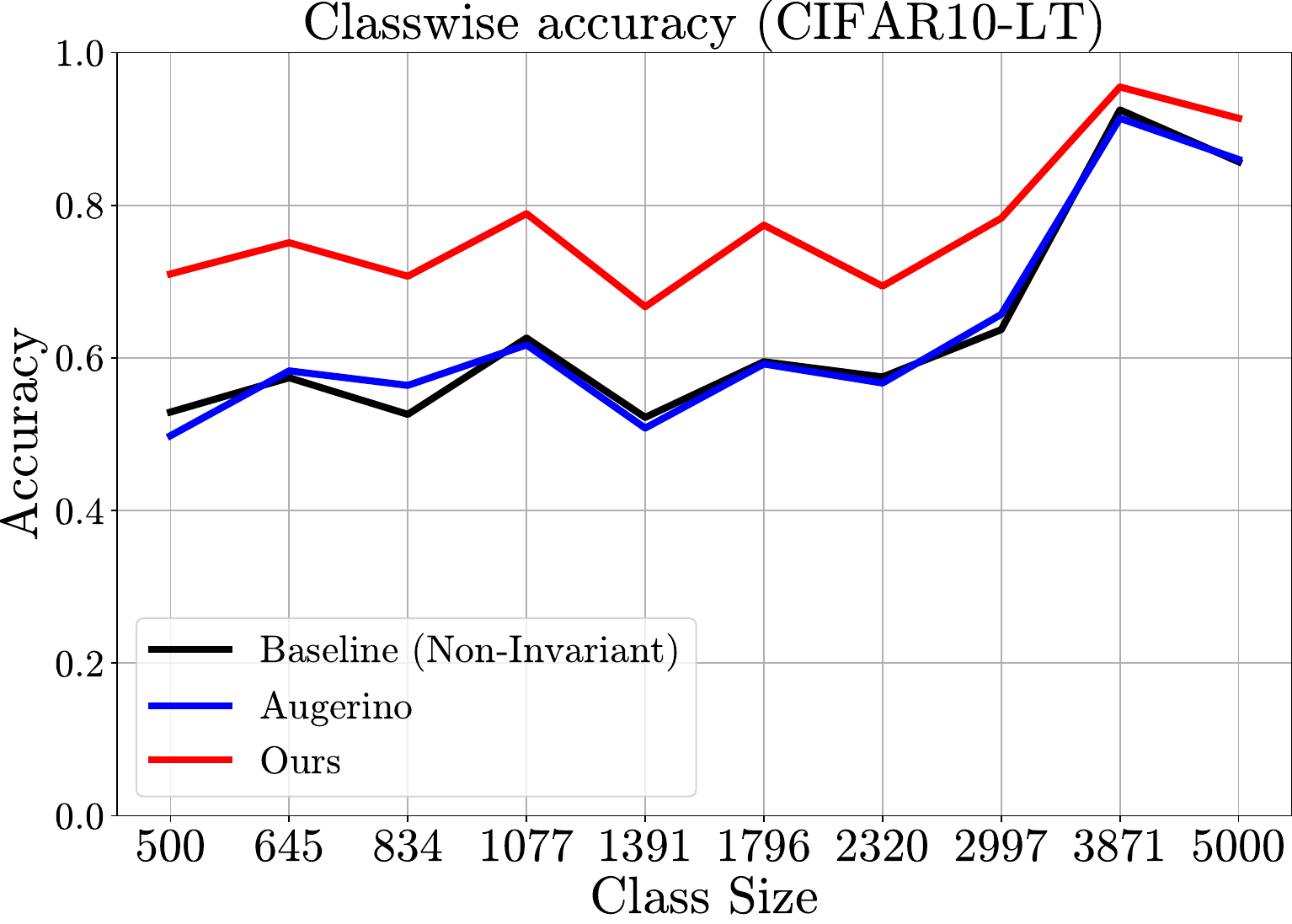}
    \caption{Our model achieves higher accuracy than Augerino and a naive model on every class on CIFAR10-LT, and the margin is larger for tail classes.}
    \label{fig:classwise}
\end{figure}

In \cref{fig:classwise}, we take the models from \cref{fig:transfer_class} for CIFAR10-LT and plot the class-wise accuracy for each. We find that our model achieves higher accuracy for each class, and the margin is larger for the tail classes.

\subsection{Mean-shift alignment failure due to multimodality}

\begin{figure}[h]
    \centering
    \includegraphics[width=0.9\linewidth]{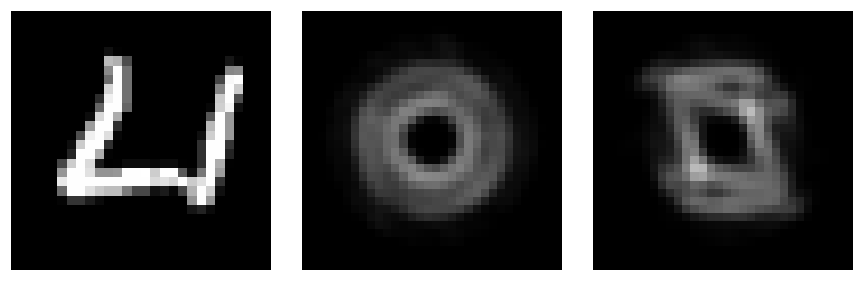} \\
    \includegraphics[width=0.9\linewidth]{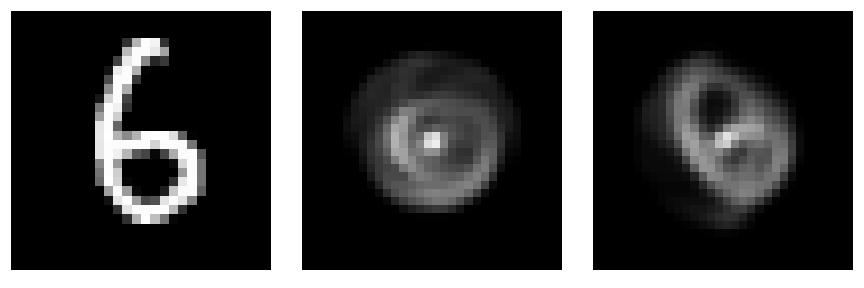} \\
    \includegraphics[width=0.9\linewidth]{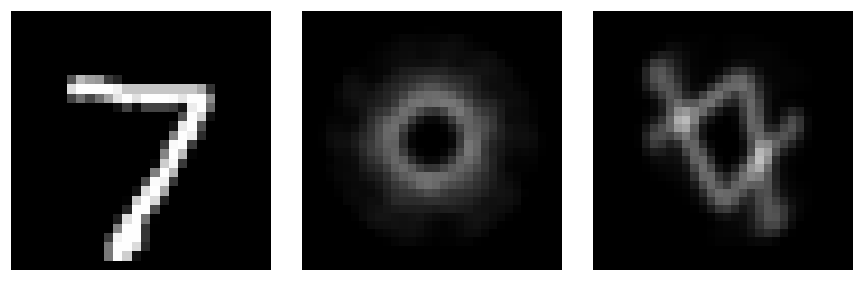} \\
    \caption{Mean-shift alignment can fail on an unseen dataset due to multi-modality. We demonstrate three examples of MNIST digits for an augmenter trained on the Mario/Iggy dataset. The digits are rotated by $\pm 180^\circ$ and processed through the mean-shift alignment algorithm. We find that for digits like $4,6,7$, the algorithm can converge to distinct modes, leading to the post-alignment images looking blurry.}
    \label{fig:meanshift_multimodality}
\end{figure}

We explore one of the reasons why our mean-shift method fails to align some digits in \cref{fig:alignment} when the augmenter is trained on Mario/Iggy and tested on MNIST. We show that this algorithm is susceptible to multiple modes as it has no information about the true pose distribution of the MNIST digits. In \cref{fig:meanshift_multimodality}, we show three examples of MNIST digits. We rotate each digit by $\pm 180^\circ$, run the alignment, and show the rotated and aligned versions superimposed. We find that for digits such as $4$, $6$, and $7$ there are multiple modes/orientations onto which the algorithm can converge. This problem is more prevalent when the variation in poses in a class is large, leading to a similar kind of blurring of the post-alignment images as observed in \cref{fig:alignment}.

\subsection{Experimental Details}
\subsubsection{Normalizing flow model} We use RealNVP \cite{dinh2017density} to model the distributions over transformation parameters with a $\operatorname{tanh}$ layer after every mixing step and at the end to ensure the model produces samples within $\big[-1, 1\big]$. The MLP at each layer of the model has $2$ layers of width $32$ each. Since the goal is an input-conditional distribution, we use a pose-embedding CNN to create an embedding vector $e$ with 32 elements. This vector is used in two places: \textbf{(1)} it is projected through a learned linear layer to the base distribution parameters (location, scale, and the weight of each mode for the conditional mixture of Gaussians), and \textbf{(2)} it is concatenated to the MLP used by RealNVP at every layer. We scale MLP and projection weights by $0.01$ to initialize them to small values, allowing the flow model to start as identity. We developed our implementation on top of normflows \cite{normflows}.

\subsubsection{Base Distribution} We use a mixture of Gaussians or uniform random variables as the base distribution. We further use the Gumbel-softmax trick \cite{gumbel} to estimate gradients with respect to the weight of each mode. We found this to be important for learning equally weighted modes in the multi-modality experiments. For experiments with simpler distributions (such as Mario-Iggy), we use a uniform distribution for simplicity.

\subsubsection{Stabilizing training with PID} A central issue in augmentation learning literature is how to use the optimal range of augmentations. If the range is too small, the model overfits, whereas if the range is too wide, the model underfits. This tradeoff is decided using a regularization constant that controls the augmentation distribution's width. Augerino fixes a regularization constant, but the resulting distribution width is difficult to control as it depends on the ratio of the classification loss and the regularization constant. This is further complicated by the fact that the classification loss changes over the course of training \cite{insta_aug}. 

InstaAug constrains the distribution \textit{entropy} to be in some pre-defined range $[H_{min}, H_{max}]$ exponentially increasing the regularization factor with each step until the entropy returns to the range. In practice, we found this method to be unstable in practice and difficult to tune. We replace this implementation with a PID controller, which adjusts the regularization constant to keep the entropy close to a set value. PID control is well understood in control systems literature and significantly easier to tune in practice. The regularization term increases linearly with the distance to the target and over time (as opposed to the exponential increase used by InstaAug), resulting in more stable dynamics.

\subsubsection{Pose-embedding CNN} We use a 5-layer CNN architecture similar to the one used by Augerino \cite{augerino}. It contains 4 convolutional layers (widths $8$, $16$, $32$, $64$ and kernel size $3$) and ReLU non-linearity after each. This is followed by max-pooling and flattening to a $64$-D vector which goes through a learned linear layer to produce the $32$-D embedding. The model has $0.03M$ parameters, which is negligible compared to the classifier for our experiments.

\subsubsection{Figure 2: Mario/Iggy experiments} We use a normalizing flow model with $4$ layers, batch size $128$, and learning rate $10^{-3}$ with the AdamW \cite{adamw} optimizer (betas: $0.9, 0.999$). We train the classifier without any augmentations for the first $5$ epochs and for $100$ epochs in total. We use the target entropy of $2$ nits, with the PID control constants $(0.01, 0.01, 0)$ (corresponding to PI control). We further smooth the inputs and outputs of the PID controller with an exponential moving average with the smoothing constant $0.9$. The base distribution is a conditional uniform distribution with $1$ mode. 

\subsubsection{MNIST (Classes $0,1,5,6,9$)} We use the same model as the previous experiment, but with $12$ layers for the normalizing flow. We optimize it for $40$ epochs with entropy regularization set of $0.3$ and trained without augmentations for the first $3$ epochs.

\subsubsection{Multi-modal experiments} We use a similar model as the MNIST classes experiment. However, we use a batch size $256$, and train without augmentations for $3$ epochs. We train for a total of $150$ epochs and use a mixture of Gaussians with $90$ modes. The Gumbel-softmax has a temperature of $0.1$. We also set the regularization factor to $0.5$ (no PID).

\subsubsection{Rotation discovery experiments in \cref{fig:distribution}} We use the same model as the multi-modal experiments. However, we use a batch size of $512$, $12$ layers, LR $1 \times 10^{-4}$ for $100$ epochs, and regularization factor to $0.002$. Finally, we use a frozen MLP trained on the Mario/Iggy dataset and the augmentation model described in Equation $23$.

 \subsubsection{Mean-shift alignment experiments} We use the model trained on the Mario/Iggy $45^\circ$, and run the mean-shift algorithm for $20$ iterations with $\alpha=0.1$ and $100$ transformation samples per iteration. For the MNIST mean-shift experiments in \cref{fig:test_time_robust}, we use $60$ iterations, $500$ samples, $\alpha=0.1$.

\subsubsection{eKLD plots} \textbf{CIFAR10-LT}: We use the trained model from CIFAR10-LT experiments, and rotate each image uniformly between $\pm 10^\circ$. We follow the method used in \cite{git}. \textbf{RotMNIST-LT}: We construct a fully rotated and long-tail version of the MNIST dataset with $\rho =10$ and train the model with batch size $128$ and $10$ modes.

\subsubsection{InstaAug experiments} We build upon the InstaAug \cite{insta_aug} codebase. Our normalizing flow model has $12$ layers. Since InstaAug already uses a CNN to convert images into a $321$-dimensional embedding, we no longer use a CNN. We project the embedding through a learned linear layer to the base distribution parameters and add the $321$-d embedding to each of the RealNVP's MLP layers. \textbf{Without LRP}: We follow the same training schedule as InstaAug, but replace their regularization scheduler with our PID ($k_p=0.1, k_i=0.5$) and their crop sampler with ours. We also set the target entropy to $2.75$. We produce samples as $4$-D vectors with entries in the $[-1,1]$ range. We then convert this vector to a crop size and crop location. To speed up the training to match InstaAug's schedule, we limit the model predictions to valid crops throughout the training. To predict valid crop sizes in the range $[\operatorname{llim}, \operatorname{ulim}]$, we translate and scale the entries accordingly, and to predict the valid crop locations, we scale the crop center prediction by $(1-size)$. The correlations between valid crop sizes and centers caused by this scaling are learned by our flow model. Initially the limits are $[0.7, 1]$, followed by the less restrictive limits $[0.35, 1]$ after $20$ epochs. \textbf{With LRP}: InstaAug's location-related parametrization (LRP) is a categorical distribution over crop locations and sizes. It already produces a good distribution of crops for TinyImageNet, so we build on it. We use InstaAug's entropy scheduler as well as crop sampler. We combine the two models by adding samples from our flow to the InstaAug output to increase crop diversity while maintaining the advantages of LRP.

\subsubsection{CIFAR10} We build upon the codebase used by LILA \cite{lila} and use their ResNet $8-16$ model. We use a regularization factor $0.1$ and initial LR $0.1$ with the augmenter LR $10^{-4}$. We train the model for $200$ epochs and reduce the LR by a factor of $10$ after every $80$ epochs. \textbf{FMNIST}: We use regularization factor $0.03$ and LR $0.03$ and augmenter LR $10^{-3}$. We train the model for $250$ epochs, reducing the LR by a factor of $20$ every $80$ epochs. For MNIST, we reduce the LR to $1e-3$ and decay it by $10$ every $20$ epochs. \textbf{CIFAR10-LT}: We use the same hyperparameters as the CIFAR10 experiment but instead reduce the LR by a factor of $20$ instead and run for $100$ epochs only and use regularization of $0.05$. \textbf{Rejection sampling for test-time augmentation}: When the augmentation budget for TTA is small (e.g. $\leq 30$ augmentations), we use rejection sampling to select the most useful augmentations. Specifically, we choose samples $w$ such that $\| w \|_2 < 1$. This step significantly reduces the variance of the output (and thus the augmentations needed), while  $\| w \|_2 < 1$ still covers the full range for any individual transformation type (i.e. $[-1, 1]$ along the axis).

\subsection{Errata} In the original draft of this paper, we included a plot using mean-shift alignment for a CIFAR-10 model for added robustness to out-of-distribution (\cref{fig:test_time_robust}). We found that the code for this plot had a bug that double-counted the $\pi$ factor in the augmentation range, resulting in higher accuracy across input rotations. As the increase in robustness could not be explained by mean-shift alignment, and since mean-shift alignment does not reliably work for CIFAR10, we have removed the plot.

\end{document}